%% file: paper.tex
\documentclass[journal]{IEEEtran}
\usepackage{cite}
\ifCLASSINFOpdf
   \usepackage[pdftex]{graphicx}
   \graphicspath{{../pdf/}{../jpeg/}}
   \DeclareGraphicsExtensions{.pdf,.jpeg,.png}
\else
\fi
\usepackage{amsmath,amsfonts,amssymb,booktabs}
\usepackage{algorithm}
\usepackage{algorithmic}
\usepackage{array,multirow}
\ifCLASSOPTIONcompsoc
  \usepackage[caption=false,font=normalsize,labelfont=sf,textfont=sf]{subfig}
\else
  \usepackage[caption=false,font=footnotesize]{subfig}
\fi
\usepackage{fixltx2e}
\usepackage{stfloats}
\usepackage{url}
\begin{document}
\title{Adaptive Fusion Affinity Graph with Noise-free Online Low-rank Representation for Natural Image Segmentation}

\author{Yang~Zhang, Moyun~Liu, Huiming~Zhang, Guodong~Sun, and Jingwu~He

\thanks{Y. Zhang and G. Sun are with the School of Mechanical Engineering, 
Hubei University of Technology, Wuhan 430068, China (e-mail: yzhangcst@hbut.edu.cn; sgdeagle@163.com).}
\thanks{M. Liu is with the School of Mechanical Science and Engineering,
Huazhong University of Science and Technology, Wuhan 430074, China (email: lmomoy8@gmail.com).}
\thanks{Y. Zhang, H. Zhang, and J. He are with the National Key Laboratory for Novel Software Technology, 
Nanjing University, Nanjing 210023, China (email: zhanghmcst@163.com; hejw005@gmail.com).}
}


\markboth{Journal of \LaTeX\ Class Files,~Vol.~00, No.~0, October~2021}%
{Zhang \MakeLowercase{\textit{et al.}}: Adaptive Fusion Affinity Graph with Noise-free Online Low-rank Representation for Natural Image Segmentation}



\maketitle

\begin{abstract}
Affinity graph-based segmentation methods have become a major trend in computer vision. The performance of these methods rely on the constructed affinity graph, with particular emphasis on the neighborhood topology and pairwise affinities among superpixels. Due to the advantages of assimilating different graphs, a multi-scale fusion graph has a better performance than a single graph with single-scale. However, these methods ignore the noise from images which influence the accuracy of pairwise similarities. Multi-scale combinatorial grouping and graph fusion also generate a higher computational complexity. In this paper, we propose an adaptive fusion affinity graph (AFA-graph) with noise-free low-rank representation in an online manner for natural image segmentation. An input image is first over-segmented into superpixels at different scales and then filtered by the proposed improved kernel density estimation method. Moreover, we select global nodes of these superpixels on the basis of their subspace-preserving presentation, which reveals the feature distribution of superpixels exactly. To reduce time complexity while improving performance, a sparse representation of global nodes based on noise-free online low-rank representation is used to obtain a global graph at each scale. The global graph is finally used to update a local graph which is built upon all superpixels at each scale. Experimental results on the BSD300, BSD500, MSRC, SBD, and PASCAL VOC show the effectiveness of AFA-graph in comparison with state-of-the-art approaches. The code is available at \url{https://github.com/Yangzhangcst/AFA-graph}.
\end{abstract}


% \fi
\IEEEpeerreviewmaketitle

\input{files/INTRODUCTION}
\input{files/RELATEDWORKS}
\input{files/METHOD}
\input{files/EXPERIMENTS}
\input{files/CONCLUSION}

\ifCLASSOPTIONcaptionsoff
  \newpage
\fi



\bibliographystyle{IEEEtran}
\bibliography{refs}



\vfill


\end{document}

%% file: files/INTRODUCTION.tex
\section{Introduction}
\label{sec:intro}
\IEEEPARstart{A}s a fundamental and challenging task in computer vision, image segmentation is a process of decomposing an image into independent regions, which plays an important role in many high-level applications~\cite{zhu2016beyond}. Unsupervised methods receive much attention because they require no prior knowledge. In the literature, many unsupervised segmentation methods have been intensively studied~\cite{li2018iterative}. Some representative works of them rely on constructing a reliable affinity graph for the representation of image content, such as the adjacency-graph~\cite{li2012segmentation}, $\ell$$_0$-graph~\cite{wang2013graph}, and global/local affinity graph (GL-graph)~\cite{wang2015global}.

\begin{figure}[!t]
	\centering
	\subfloat[]{
		\includegraphics[width=1.1in]{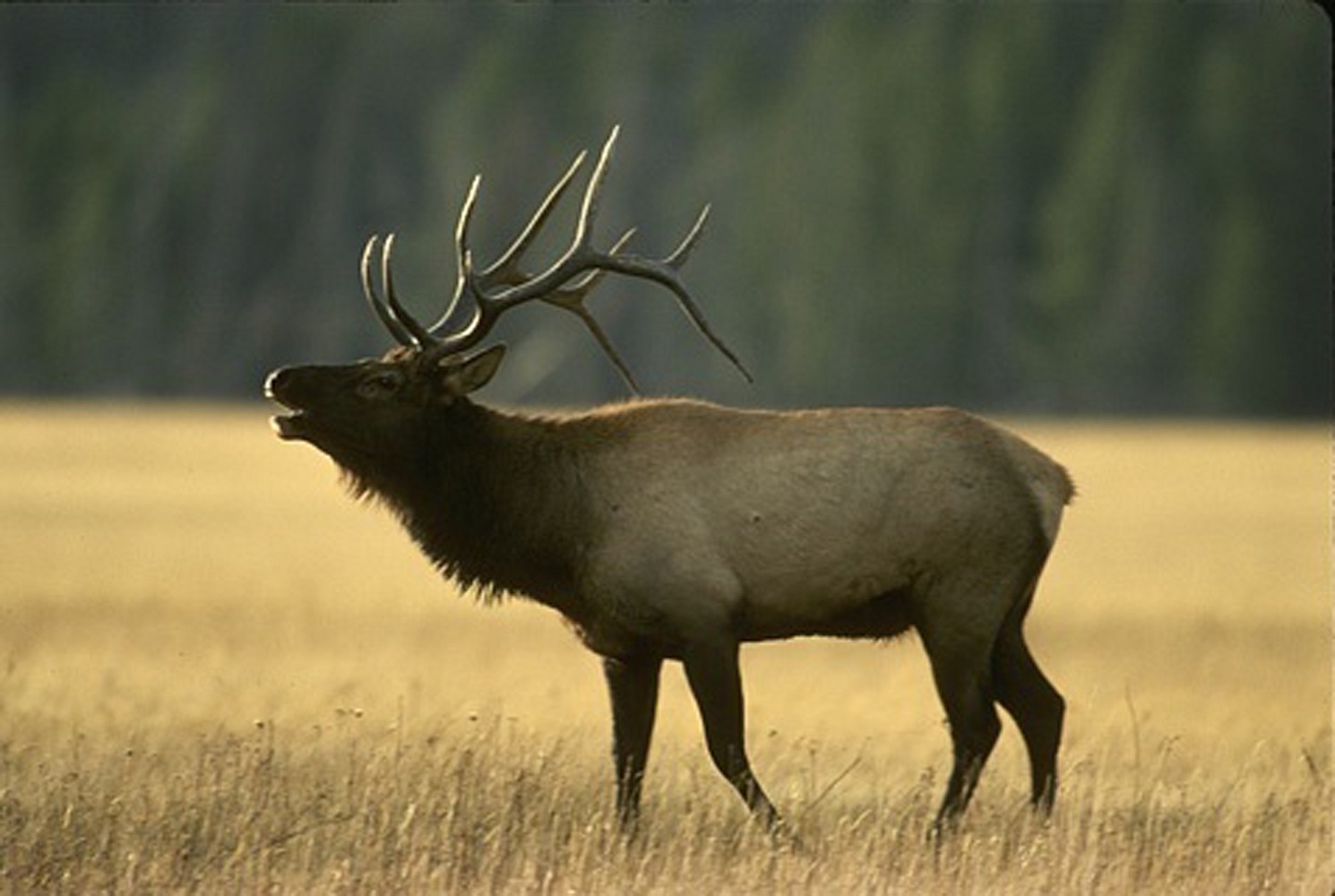}
	}\hspace{-0.8em}
	\subfloat[]{
		\includegraphics[width=1.1in]{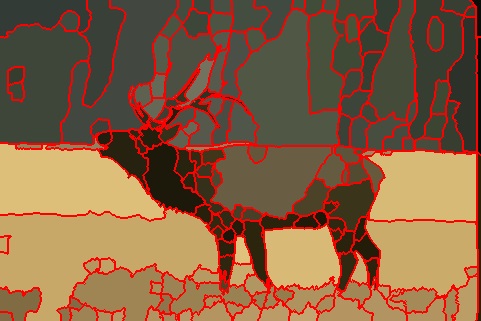}
	}\hspace{-0.8em}
	\subfloat[]{
		\includegraphics[width=1.1in]{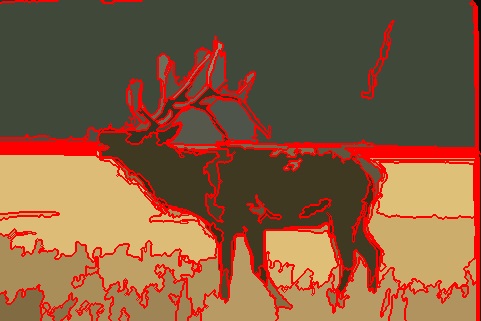}
	}
	\vspace{-0.5em}

	\subfloat[]{
		\includegraphics[width=1.1in]{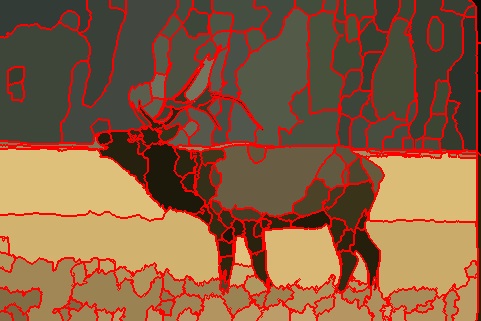}
	}\hspace{-0.8em}
	\subfloat[]{
		\includegraphics[width=1.1in]{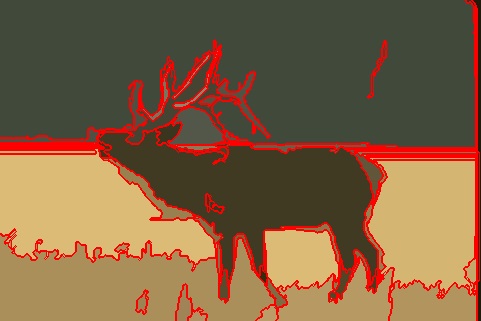}
	}\hspace{-0.8em}
	\subfloat[]{
		\includegraphics[width=1.1in]{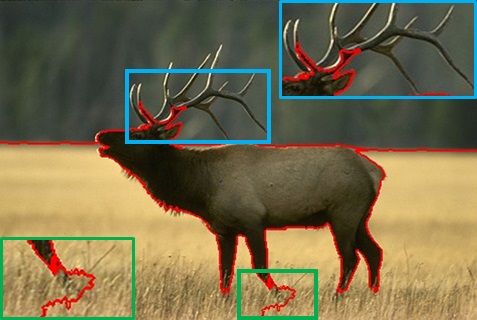}
	}
	\vspace{-0.5em}

	\subfloat[]{
		\includegraphics[width=1.1in]{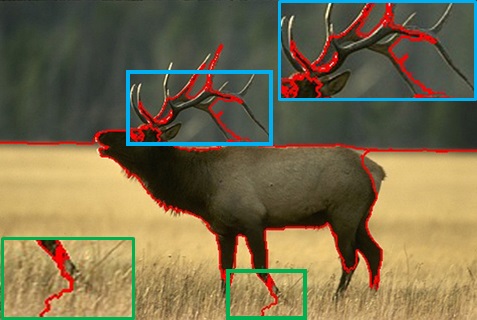}
	}\hspace{-0.8em}
	\subfloat[]{
		\includegraphics[width=1.1in]{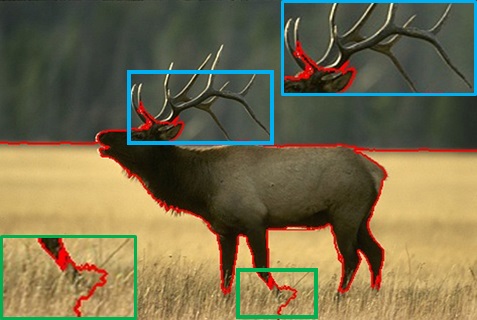}
	}\hspace{-0.8em}
	\subfloat[]{
		\includegraphics[width=1.1in]{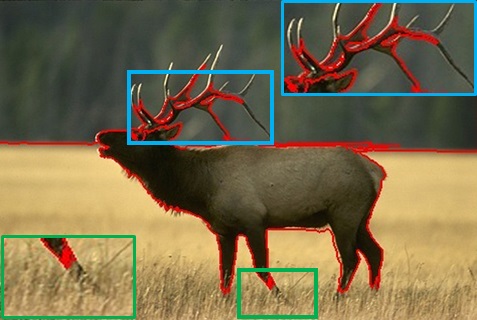}
	}
	\caption{Comparison results by different graph-based segmentation methods. (a) Input image. (b-e) Superpixels generated by over-segmenting the image at different scales. (f-i) Segmentation results by adjacency-graph~\cite{li2012segmentation}, $\ell_0$-graph~\cite{wang2013graph}, GL-graph~\cite{wang2015global}, and our AFA-graph, respectively. Although superpixels features are extremely complex and vary greatly at different scales, our method enforces the global structure  and preserves more local information simultaneously.}
\label{introfig}
\end{figure}

Clearly, for these affinity graph-based methods, the segmentation performance significantly depends on the effectiveness of the constructed affinity graph, with particular emphasis on the neighborhood topology and pairwise affinities between nodes (\textit{i.e.}, pixels or superpixels). For example, adjacency-graph~\cite{li2012segmentation} of coherent superpixels is applied to aggregate cues for segmentation. However, it has not enough ability to collect global grouping cues when the objects occupy a large of areas in the image. $\ell_0$-graph~\cite{wang2013graph} uses sparse representation, which enables it to capture global grouping cues, but it has tendency to create isolated regions. As shown in Fig. 1, GL-graph~\cite{wang2015global} is proposed to combine the adjacency-graph and the $\ell_0$-graph according to superpixel areas at different scales, leading to a better result than only using a single graph. The superpixels are simply divided into three parts: small, medium, and large sized according to their area in GL-graph~\cite{wang2015global}. However, there still exist three difficulties to be solved: i) these affinity graph-based methods ignore the noise in images and features, which can influence the accuracy of pairwise similarities and further hinder the affinity propagation; ii) due to the extreme complexity of superpixel features and the wide variation at different scales, the combination principles of different graphs are unreliable and often rely on empiricism; iii) multi-scale combinatorial grouping with graph construction are obliged to generate additional calculations, incurring a higher computational complexity.

To solve these problems, an \emph{adaptive fusion affinity graph} (AFA-graph) is proposed to segment natural images. The proposed AFA-graph combines the local and global nodes of superpixels at different scales based on affinity propagation. The superpixels are obtained by over-segmenting an input image at different scales, and then filtered by an improved kernel density estimation method. We use affinity propagation clustering to select global nodes of these superpixels according to their subspace-preserving representation. Moreover, the noise-free online low-rank representation (NOLRR) is used to obtain a NOLRR-graph at each scale via a sparse representation of mLab features of the global nodes. All superpixels at each scale are used to build an adjacency-graph, which are further updated by the NOLRR-graph. We introduce a bipartite graph to map the relationship between the original image pixels and superpixels, and to enable propagation of grouping cues across superpixels at different scales. Intensive experiments are conducted on five public datasets, namely BSD300, BSD500, MSRC, SBD, and PASCAL VOC with four metrics, including PRI, VoI, GCE, and BDE for quantitative comparisons. The experimental results show the effectiveness of our AFA-graph compared with the state-of-the-art methods.

This work makes the following contributions.

\begin{itemize}
  \item We explore the impact of noise on pairwise similarity and affinity propagation between superpixels.
  \item We construct an AFA-graph to combine different graphs with the sparsity and a high discriminating power for natural image segmentation.
  \item We propose a novel NOLRR-graph to improve segmentation performance while reducing the time complexity.
  \item Extensive quantitative and qualitative experiments show that our AFA-graph outperforms recent methods on five benchmark datasets.
\end{itemize}

A preliminary conference version of this paper can be referred to adaptive affinity graph with subspace pursuit (AASP-graph)~\cite{zhang2019aaspgraph}. Compared with AASP-graph~\cite{zhang2019aaspgraph}, this study contains: i) an adaptive fusion affinity graph named as AFA-graph is proposed to segment natural images with online low-rank representation; ii) we propose a novel global node selection strategy containing subspace-preserving representation and affinity propagation clustering. As a new algorithmic enhancement, the strategy further perfects global node selection and improves the segmentation performance; iii) we propose a NOLRR-graph to improve segmentation accuracy and reduce computational complexity; iv) we perform extensive experiments to validate the effectiveness of global node selection strategy and the NOLRR-graph. More extra experiments on five datasets and theoretical analyses are also added to verify the robustness of AFA-graph. The official code is available at \url{https://github.com/Yangzhangcst/AFA-graph}.

The rest of this paper is organized as follows. Related works are reviewed in Section II. The proposed AFA-graph for natural image segmentation is presented in Section III. Experimental results are reported in Section IV, and the paper is concluded in Section V.

%% file: files/RELATEDWORKS.tex
\section{Related Works}
The unsupervised methods segment images without any human intervention. As we focus on unsupervised image segmentation in this paper, a review of related methods is introduced in this section. A more detailed review of the image segmentation process can be found in~\cite{zhu2016beyond}.

The essence of image segmentation can be regarded as a clustering problem, which groups the pixels into local homogenous regions. Some clustering-based methods such as $k$-means, mean-shift (MS)~\cite{comaniciu2002mean},  fusion of clustering results (FCR)~\cite{4480125}, correlation clustering (Corr-Cluster)~\cite{Kim2013Task}, higher-order correlation clustering (HO-CC)~\cite{nowozin2014image}, deviation-sparse fuzzy c-means (DSFCM)~\cite{8543645}, membership scaling fuzzy c-means (MSFCM)~\cite{9120181}, FRFCM~\cite{8265186}, automatic fuzzy clustering (AFC)~\cite{8770118}, robust self-sparse fuzzy clustering (RSFFC)~\cite{9162644} and superpixel-based fast fuzzy c-means (SFFCM)~\cite{Lei2018fuzzy} are the typical examples. In particular, subspace clustering~\cite{you2016scalable,feng2013online} has extensively established in computer vision. Many subspace clustering algorithms aim to obtain a structured representation to fit the underlying data, such as sparse subspace clustering~\cite{elhamifar2013sparse} and LRR~\cite{liu2010robust}. Both of them utilize the idea of self-expressiveness which expresses each sample as a linear combination of the remaining~\cite{you2016scalable}. However, such approaches can be computationally expensive with high memory cost. One of the most popular ways to alleviate the computational complexity is online implementation~\cite{feng2013online,shen2016online}.

Besides clustering-based methods, graph-based methods can be regarded as image perceptual grouping and organization methods which have become one of the most popular image segmentation methods. Graph-based methods are based on the fusion of the feature and spatial information, such as normalized cut (Ncut)~\cite{shi2000normalized}, Felzenszwalb-Huttenlocher (FH) graph-based method~\cite{Felzenszwalb2004Efficient}, $\ell_0$-graph~\cite{wang2013graph}, iterative ensemble Ncut~\cite{Li2016Iterative}, multi-scale Ncut (MNcut)~\cite{cour2005spectral}, GL-graph~\cite{wang2015global}, region adjacency graph (RAG)~\cite{7484679}, and AASP-graph~\cite{zhang2019aaspgraph}, etc. More recently, Kim et al.~\cite{Tae2013Learning} proposed a multi-layer sparsely connected graph to effectively combine local grouping cues, and then applied semi-supervised learning to define the relevance scores between all pairs of these graph nodes as the affinities of the graph. Wang et al.~\cite{wang2014regularized} introduced the normalized tree partitioning and the average tree partitioning to optimize normalized and average cut over a tree for image segmentation. Saglam and Baykan~\cite{Saglam2017Sequential} used Prim's sequential representation of the minimum spanning tree for image segmentation. Especially, Li et al.~\cite{li2018iterative} proposed a region adjacency cracking method to adaptively loosen the color labeling constraints.

Furthermore, a heuristic four-color labeling algorithm was used to establish a uniform appearance of those regions with global consistency. In this method, affinity propagation clustering~\cite{frey2007clustering} is applied to crack the adjacent constraint, which is suitable for the task with unknown regional characteristic distribution. Affinity propagation can adaptively determine the clusters based on the relationship of the given regions. The GL-graph~\cite{wang2015global} is built over superpixels to capture both short- and long-range grouping cues, and thereby enabling propagation of grouping cues between superpixels of different scales based on a bipartite graph~\cite{li2012segmentation}. It can be noticed that the combination of the global graph and the local graph over superpixels can obtain a better segmentation result. 

%% file: files/METHOD.tex
\section{FRAMEWORK}
In this section, we first introduce the overview of our proposed AFA-graph. Then, we show the global nodes selection. Finally, we present the process of NOLRR-graph construction in details.
\begin{figure*}[!t]
\centering
\includegraphics[width=7in]{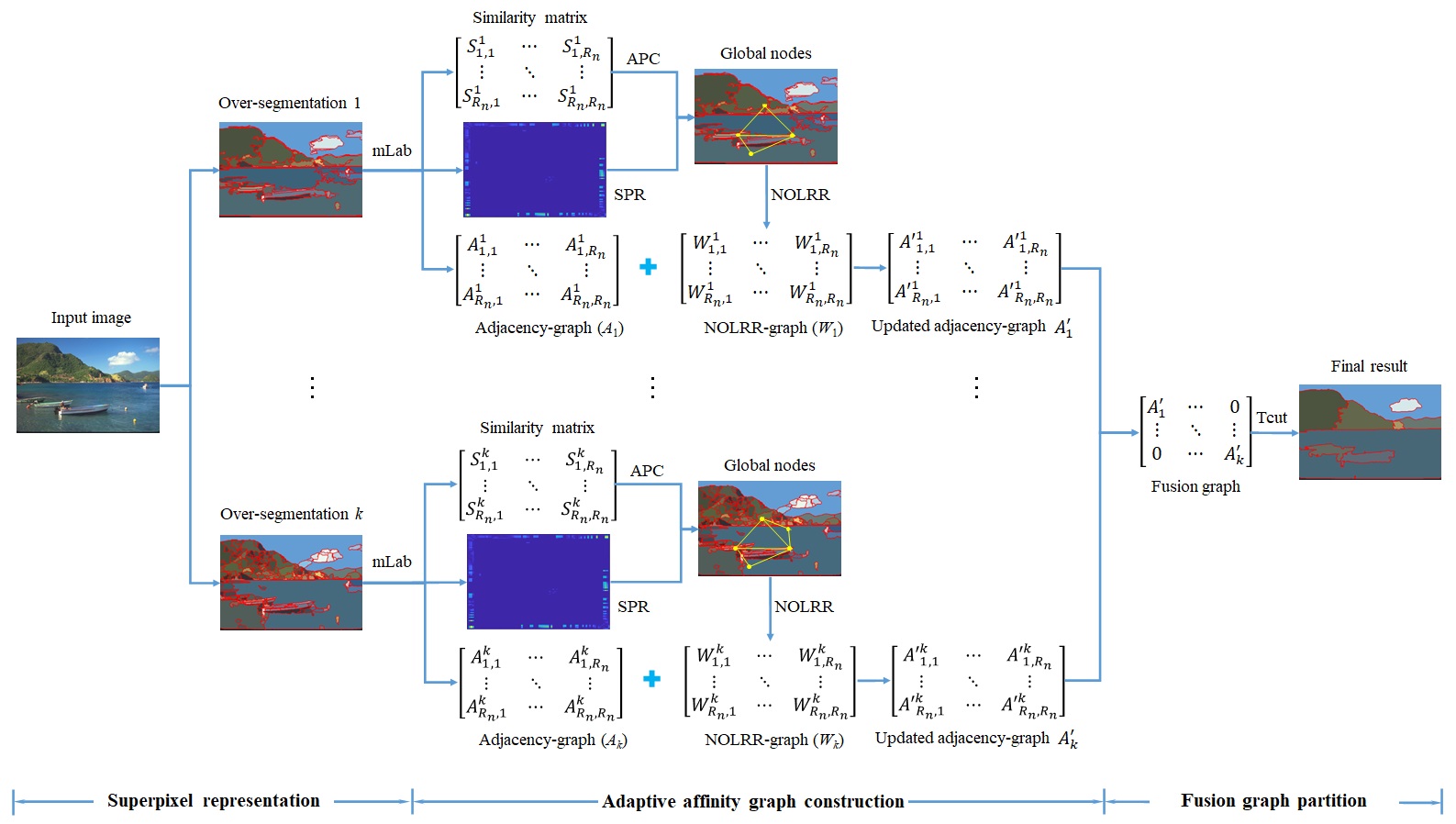}
\caption{The overall framework of image segmentation based on the proposed AFA-graph. After over-segmenting an input image as the SAS~\cite{li2012segmentation}, we obtain superpixels at $k$ different scales. Global nodes of superpixels are sorted out through subspace-preserving representation (SPR) and affinity propagation clustering (APC) and then build a NOLRR-graph at each scale. The adjacency-graph is constructed by all superpixels and it is updated by the NOLRR-graph at each scale. The updated graphs are fused to obtain the final result through Tcut~\cite{li2012segmentation}.}
\label{framework}
\end{figure*}

\subsection{Overview}
\label{overview}
The overall framework of the proposed AFA-graph for natural image segmentation is shown in Fig.~\ref{framework}. The proposed approach primarily consists of three components: superpixel representation, adaptive affinity graph construction, and fusion graph partition. For graph-based segmentation, approximating each superpixel in the feature space into a linear combination of other superpixels are regarded as neighbors, and their affinities are calculated from the corresponding representation error~\cite{wang2015global}. Let $\textbf{\emph{S}}_k=\{s\}^N_{i=1}$ be the superpixels of an input image \emph{\textbf{I}} at scale $k$, where $N$ is the number of superpixels. Formally, such an approximation can be written as:

\begin{equation}
\textbf{\emph{d}}_{j}=\textbf{\emph{D}}\textbf{\emph{c}}_{j}, \quad c_{jj}=0,
\end{equation}
where $\textbf{\emph{d}}_{j}\in\mathbb{R}^n$ is a matrix representation of superpixels over the dictionary $\textbf{\emph{D}}$, and $\textbf{\emph{c}}_{j}\in\mathbb{R}^N$ is the sparse representation of superpixels. The constraint $c_{jj}=0$ prevents the self-representation of $\textbf{\emph{d}}_{j}$.

As for all superpixels at each scale, every superpixel is connected to its adjacent superpixels denoted as local graph. Let $\textbf{\emph{M}}_A$ be the matrix-representation of its adjacent neighbors, we attempt to represent its feature $\textbf{\emph{f}}_i$ as a linear combination of elements in $\textbf{\emph{M}}_A$. In practice, we solve the following problem:
\begin{eqnarray}
\textbf{\emph{c}}''_i=\mathop{\arg\min}_{\textbf{\emph{c}}'_i}||\textbf{\emph{f}}_i-\textbf{\emph{M}}_A\textbf{\emph{c}}'_i||_2.
\end{eqnarray}
The affinities coefficient $A_{i,j}$ of the adjacency-graph $\textbf{\emph{A}}$ between superpixels $s_i$ and $s_j$ are calculated as $A_{i,j}=1$, if $i$ is equal to $j$; $A_{i,j}=1-(r_{i,j}+r_{j,i})/2$, otherwise; where $r_{i,j}=||\textbf{\emph{f}}_i-\textbf{\emph{c}}''_{i,j}\textbf{\emph{f}}_j||^2_2$.
To combine different graph, the proposed NOLRR-graph $\textbf{\emph{W}}$ is replaced by the adjacency-graph $\textbf{\emph{A}}$ at global nodes to obtain the updated adjacency-graph $\textbf{\emph{A}}'$ at a certain scale. To map the relationships between pixels and superpixels and enable propagation of grouping cues across superpixels at different scales, a bipartite graph is built to describe the relationships of pixels to superpixels and superpixels to superpixels. In this case, the bipartite graph is unbalanced, which can be solved by Tcut~\cite{li2012segmentation} to obtain the final segmentation result.

\subsection{Superpixel representation}
Superpixel is an irregular pixel block composed of adjacent pixels with similar features. It is usually used as a preprocessing segmentation step. Various visual patterns of a natural image can be captured through superpixels generated by different methods with different parameters. Like the GL-graph~\cite{wang2015global}, an input image is over-segmented into superpixels using the same parameters as in SAS~\cite{li2012segmentation}. And then, the color features of each superpixel are characterized by using mean value in the CIE L*a*b* space (mLab). 

In general, all natural images contain noise, which influences the accuracy of pairwise similarities and further hinders the affinity propagation. The disturbance of the noise to a natural image is reflected in its statistical features. And the mLab can be regarded as a special statistical feature (histogram) of the image. As pointed out by~\cite{foster2014segmentation}, a peak in the histogram corresponds to a relatively more homogeneous region in the image. The mLab of an image can be estimated in a robust state such that an estimated feature is less sensitive to local peculiarities. 

\begin{figure}[!t]
	\centering
	\subfloat[]{
		\includegraphics[width=3.3in]{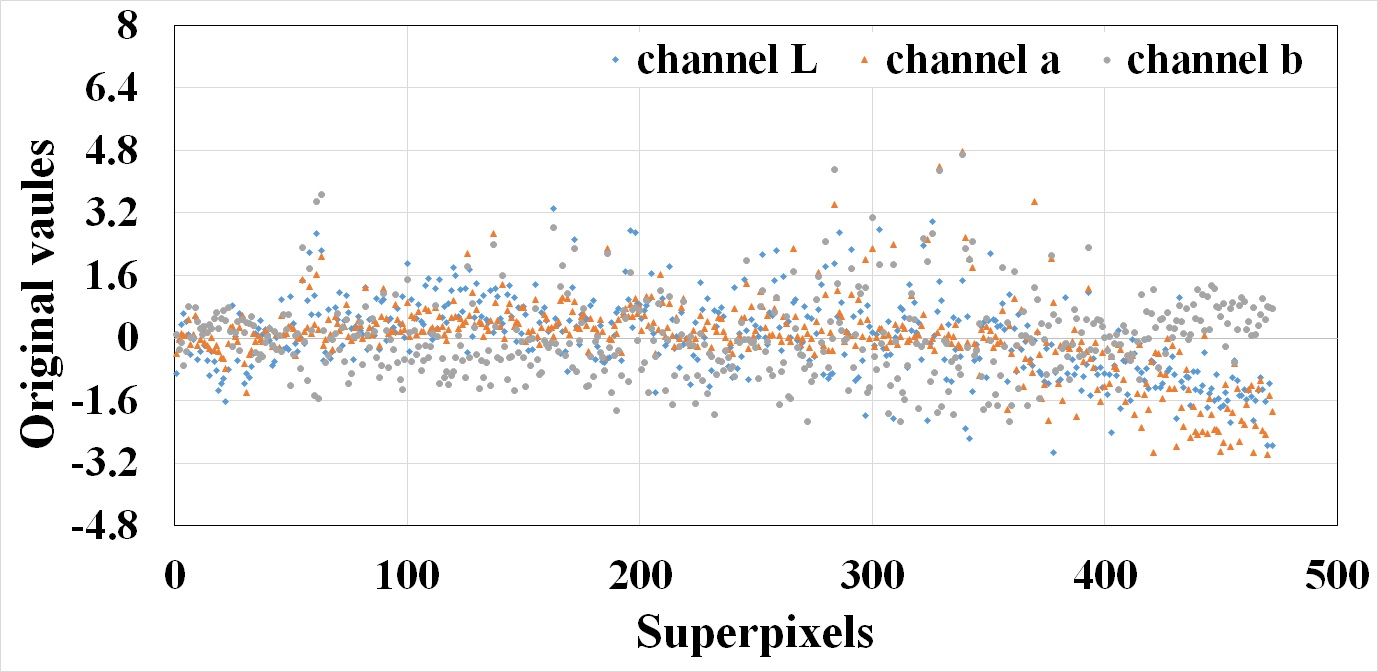}
	}
	\vspace{0.5em}
	\subfloat[]{
		\includegraphics[width=3.3in]{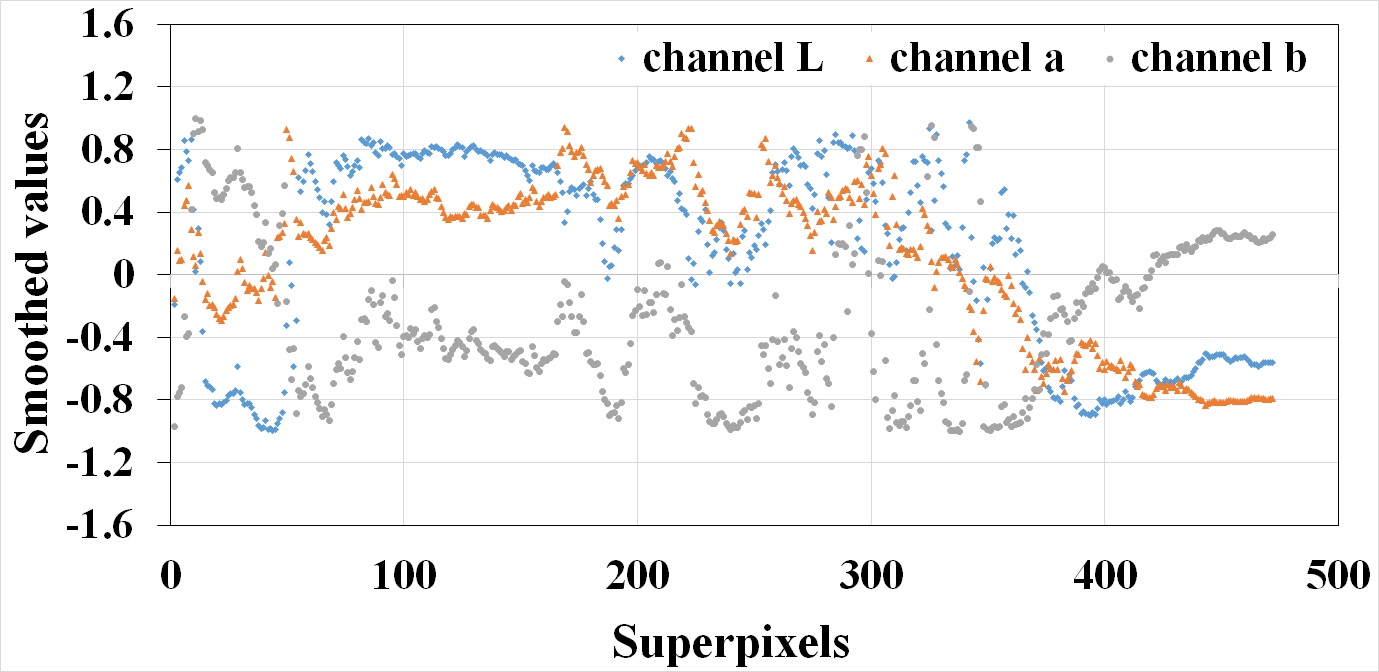}
	}
	\caption{Illustration of the IKDE with exponential smoothing method. (a) Original values of the superpixels in Lab color space. (b) Smoothed values of the superpixels processed by the IKDE.}
	\label{ikde}
\end{figure}

To reduce the noise, we introduce an improved kernel density estimation (IKDE) method to estimate the mLab of natural images, and the histogram is smoothed by single exponential smoothing. 
\begin{equation}
\textbf{\emph{f}}_{t}^{in}= \alpha \textbf{\emph{f}}_t+(1-\alpha)\textbf{\emph{f}}_{t-1}^{in}
\end{equation}
where $\alpha \in (0,1]$ is an exponential factor. $\textbf{\emph{f}}^{in}_{t}$ is the original features at time period $t$, and $\textbf{\emph{f}}_t$ is the features smoothed by IKDE. As shown in Fig.~\ref{ikde}, the essential shape of mLab features is preserved throughout this process. The IKDE can improve the reliability of features as explained in subsection~\ref{subsec:different modules}.

\subsection{Global Node Selection}
\label{node_selection}
It is shown in~\cite{you2016scalable} that superpixels from different groups can be well approximated by a union of low dimensional subspaces. One way to achieve this goal is to use the sparse subspace clustering~\cite{elhamifar2013sparse} which separates superpixels into groups such that each group contains only superpixels from the same subspace. Therefore, global nodes can be classified according to their subspace-preserving representation (SPR) which is the affiliation $\textbf{\emph{c}}_{j}$ of superpixels with subspace. It is proved that the sparsest solution of $\textbf{\emph{c}}_{j}$ measured in the sense of $\ell_0$-norm is unique, and it conveys the most meaningful information of superpixels~\cite{MichaelElad10}. So, the sparse solution can be regarded as follows:
\begin{equation}
\textbf{\emph{c}}_{j}^{*}= \mathop{\arg\min}_{\textbf{\emph{c}}_j}\|\textbf{\emph{c}}_{j}\|_0 \quad s.t.\ \textbf{\emph{f}}_j=\textbf{\emph{F}}\textbf{\emph{c}}_j, \ c_{jj}=0,
\end{equation}
where $\|\cdot\|_0$ represents the $\ell$$_0$-norm which counts the number of nonzero values in a vector, and $\textbf{\emph{F}}=[\ \emph{\textbf{f}}_1,...,\textbf{\emph{f}}_N]\in \mathbb{R}^{n\times N}$ is the smoothed feature matrix.

\begin{figure*}[!t]
	\centering
	\subfloat[]{
		\includegraphics[width=2.24in]{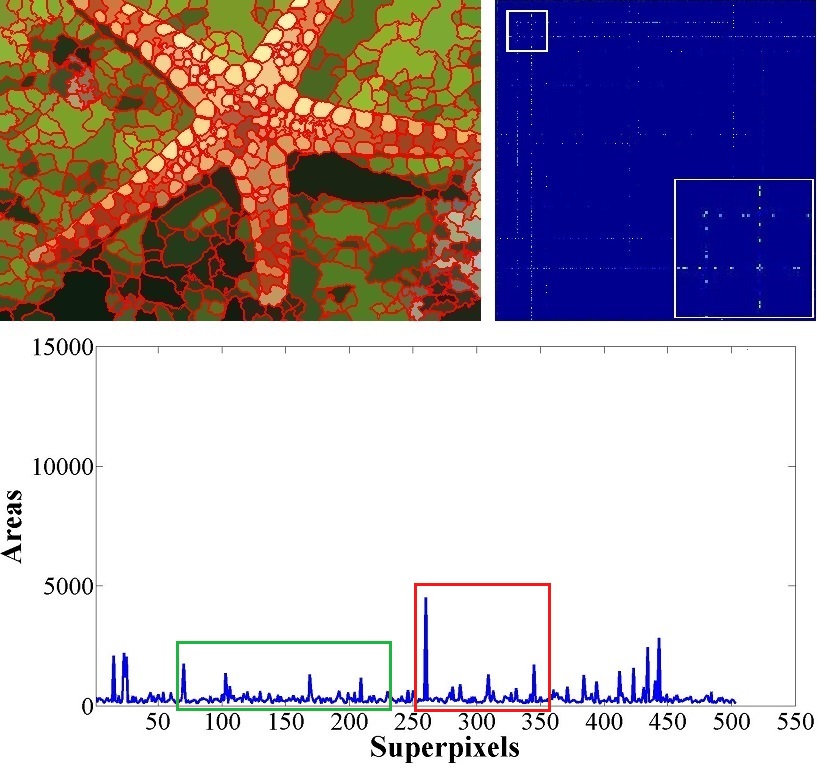}}
	\subfloat[]{
		\includegraphics[width=2.24in]{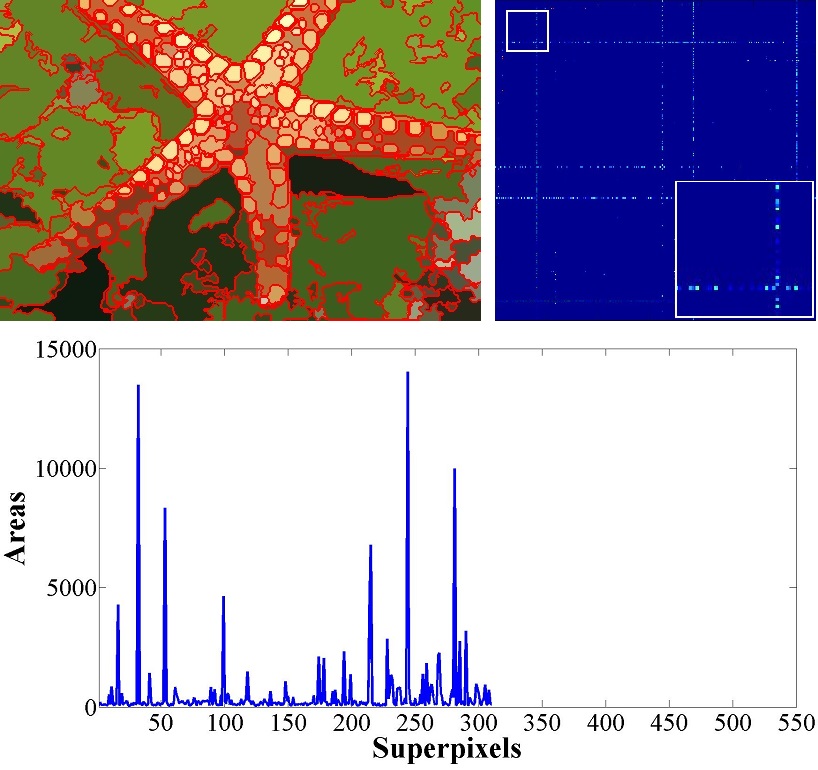}}
	\subfloat[]{
		\includegraphics[width=2.24in]{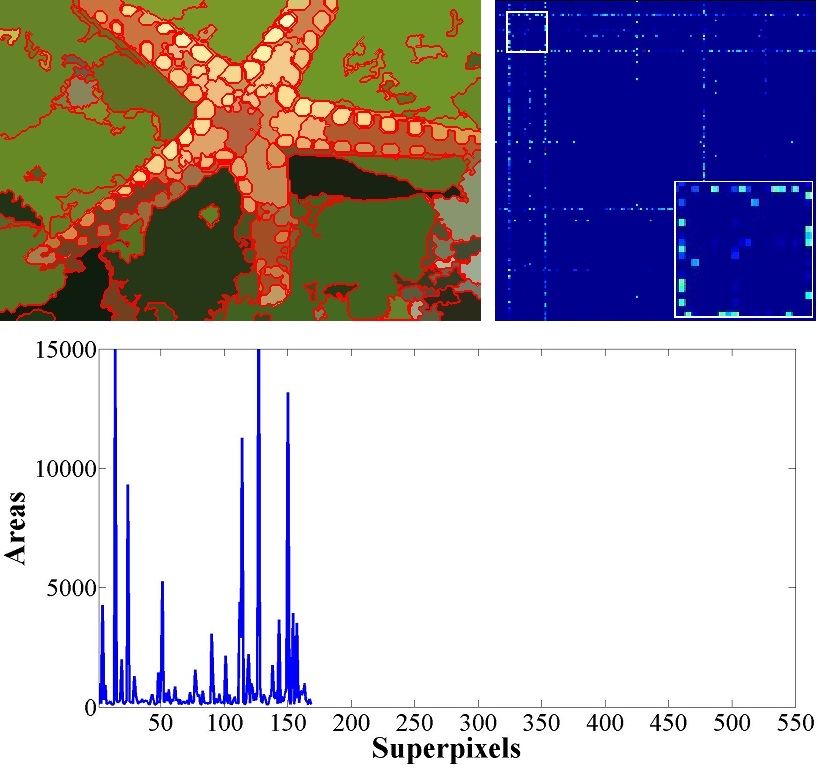}}
	\vspace{-0.5em}
	\subfloat[]{
		\includegraphics[width=2.24in]{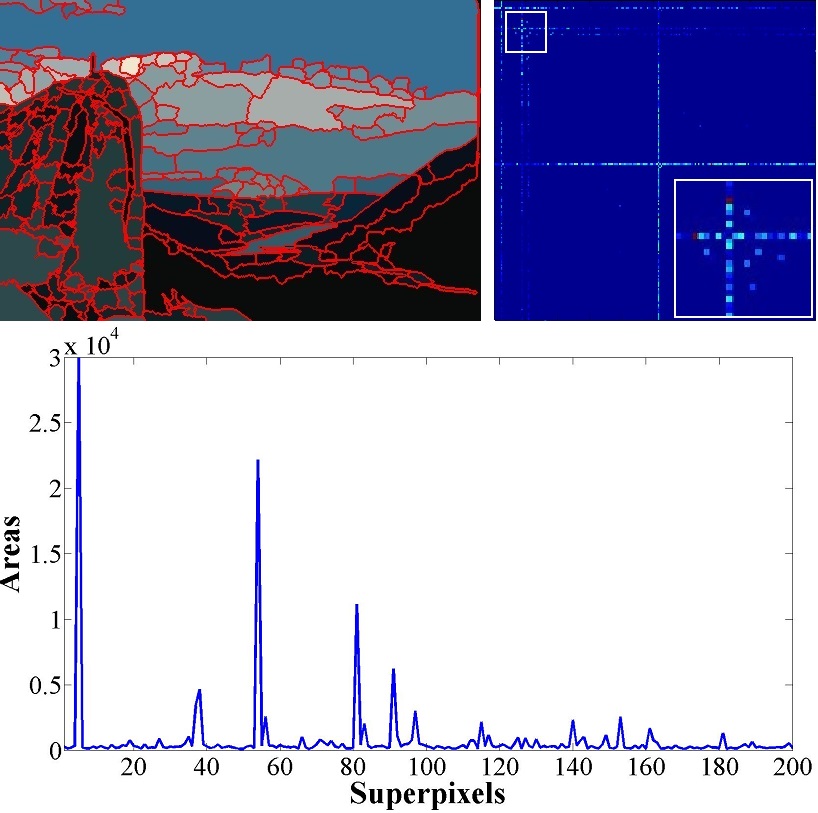}}
	\subfloat[]{
		\includegraphics[width=2.24in]{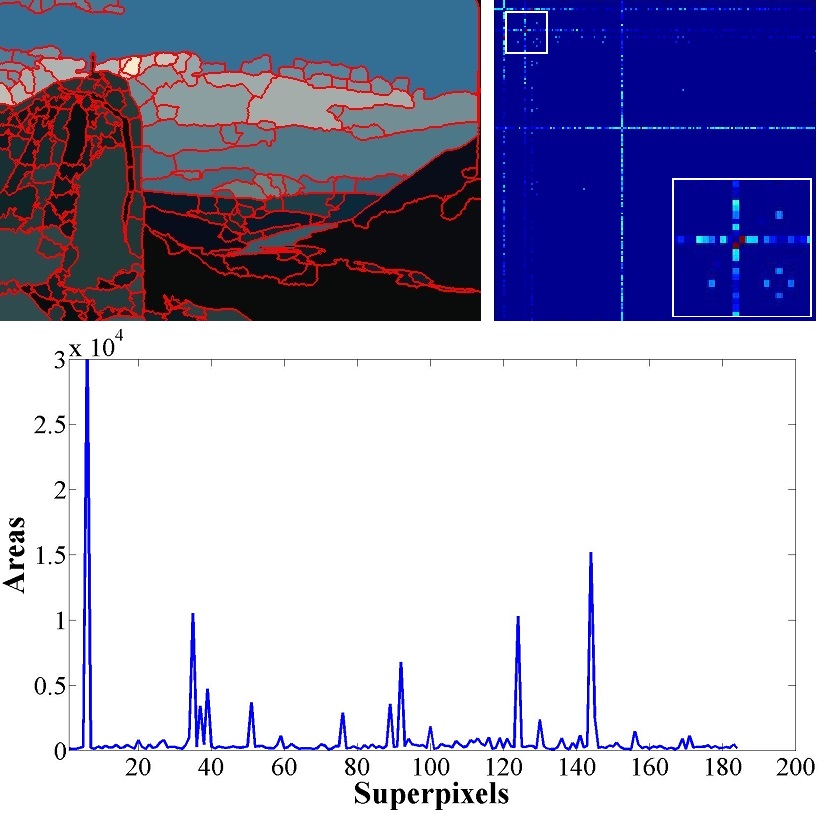}}
	\subfloat[]{
		\includegraphics[width=2.24in]{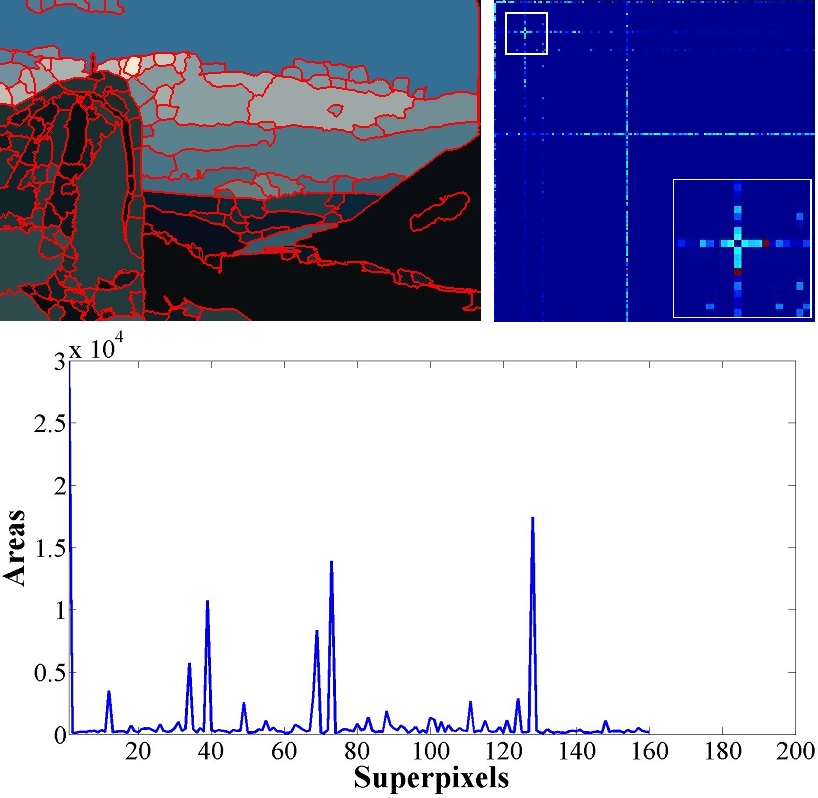}}
	\caption{The \emph{Superpixels} (top left) generated by over-segmenting the input images at different scales with their \emph{SPR} (top right) computed by mLab features and their \emph{Areas} (bottom). From left to right, the number of superpixels in the image decreases gradually. The change of superpixels has less effect on SPR than on areas in the same image.} %The SPR of different images vary significantly, but the distribution of areas (red box in (a) and (d), green box in (a) and (e)) may change a little.}
	\label{SPR}
\end{figure*}

Hence, the orthogonal matching pursuit (OMP)~\cite{you2016scalable} method is applied to seek an approximation of the sparsest solution.
\begin{equation}
\textbf{\emph{c}}_{j}^{*}= \mathop{\arg\min}_{\textbf{\emph{c}}_j}\|\textbf{\emph{f}}_j-\textbf{\emph{F}}\textbf{\emph{c}}_j\|_2^2 \quad s.t.\ \|\textbf{\emph{c}}_{j}\|_0\leq \psi, \ c_{jj}=0,
\end{equation}
where the parameter $\psi$ is the maximal number of coefficients for each input smoothed feature $\textbf{\emph{f}}_j$, which controls the sparsity of the solution. The solution $\textbf{\emph{c}}_j^{*}\in \mathbb{R}^{N}$ (the \emph{j}-th column of $\textbf{\emph{C}}^{*}\in \mathbb{R}^{N\times N}$) is computed by OMP$(\textbf{\emph{F}}_{-j}, \textbf{\emph{f}}_j)\in \mathbb{R}^{N-1}$ with a zero inserted in its \emph{j}-th entry, where $\textbf{\emph{F}}_{-j}$ is the smoothed feature matrix with the \emph{j}-th column removed.
To better illustrate the relationship between features and SPR of superpixels, we show superpixels generated by over-segmenting two images at different scales and compare the SPR of superpixels computed by mLab with the areas of superpixels in Fig.~\ref{SPR}. It can be seen that the influence of superpixels across different scales on the SPR is smaller than that of the area in the same image. Therefore, the SPR can better represent superpixels compared with the area.

\begin{figure}[!t]
	\centering
	\subfloat[]{
		\includegraphics[width=1.8in]{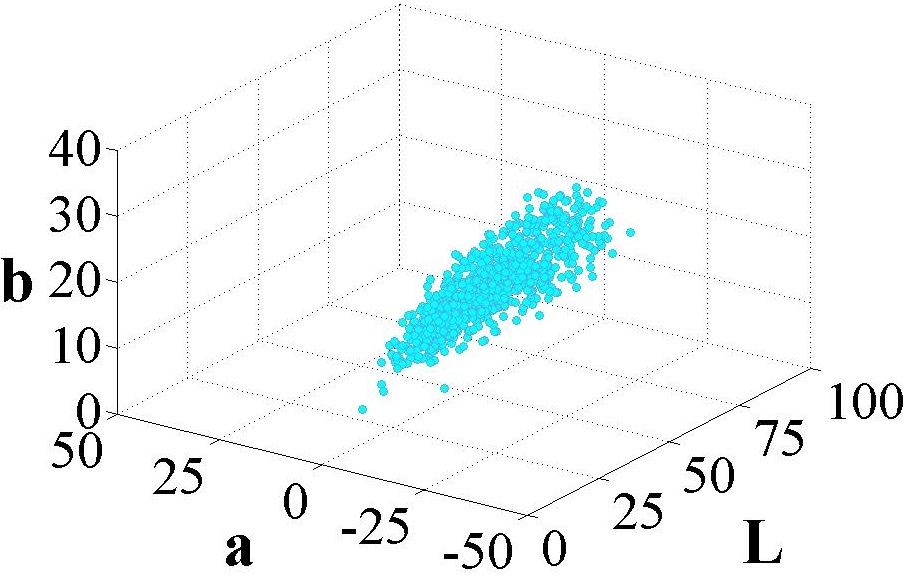}}
	
	\vspace{-0.5em}
	\subfloat[]{
		\includegraphics[width=3in]{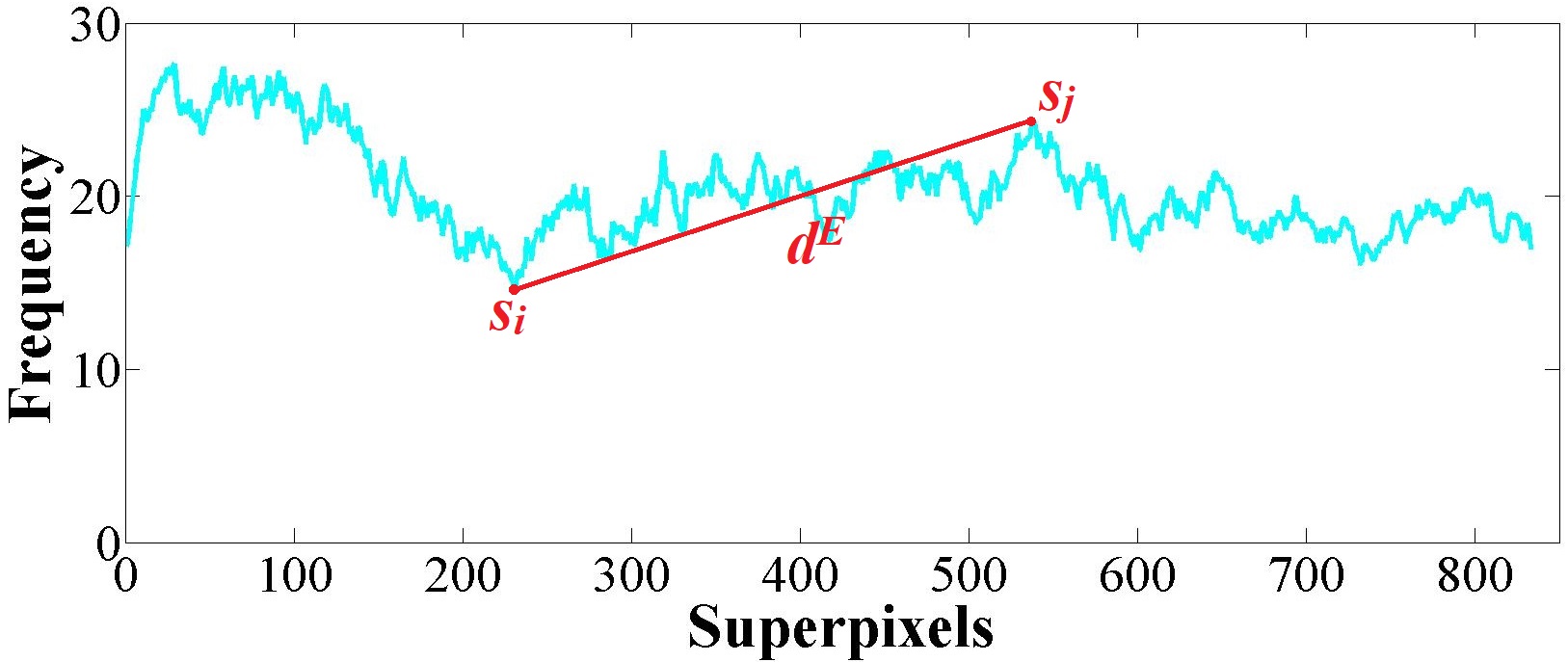}}
	\caption{Illustration of the similarity $S_{i,j}$ between two superpixels $s_i$ and $s_j$. (a) Visualization results of mLab features. (b) Euclidean distance along the probability density function of the mLab feature histogram.}
	\label{similarity}
\end{figure}

In Fig.~\ref{SPR}, the features of superpixels are various at different scales. The global nodes cannot be easily defined in natural images. To address this issue, we propose a novel similarity between any two superpixels and apply affinity propagation clustering (APC)~\cite{frey2007clustering} with the similarity matrix to find global nodes. As shown in Fig.~\ref{similarity}, the proposed similarity $S_{i,j}$ between two superpixels $s_i$ and $s_j$ is defined as a combination of the Euclidean distance $d^E_{ij}$ and the geodesic distance along the probability density function of their feature histogram.
\begin{equation}
S_{i,j} = -\Big(|d^E_{ij}|^e+|\sum_{x=i}^{j-1}d^E(x,x+1)|^g\Big)^{1/2},\ j>i.
\end{equation}
where $e$ and $g$ are the hyper-parameters which are setting as $e=3$ and $g=5$, respectively. It can be noticed that APC is very suitable for the task that we know nothing about the feature distribution of the superpixels beforehand. After $\textbf{\emph{c}}_j^{*}$ is computed, the classification of the superpixels is obtained by applying spectral clustering to the affinity matrix $\textbf{\emph{M}}_{sp}=|\textbf{\emph{C}}^{*}|+|{\textbf{\emph{C}}^{*}}^\mathsf{T}|$ with the clustering number provided by the APC. The procedure of global node selection is summarized in \textbf{Algorithm~\ref{alg:global nodes selection}}. We will give a more detailed discussion about the reliability of SPR and APC in Section~\ref{subsec:different modules}.

\subsection{NOLRR-graph Construction} 
\label{graph_construction}
To capture long-range grouping cues, we utilize Eq. (1) to approximate each global node from others in mLab features. As an alternative, we choose low-rank minimization to approximate each global node from others in mLab features. LRR is a significant method for segmenting data that are generated from a subspace union~\cite{liu2010robust,lu2016face}. Both SSC and LRR utilize the idea of expressing each sample as a linear combination of the remaining. So, it solves the following problem:
\begin{equation}
\mathop{\min}_{\textbf{\emph{C}}}{\dfrac{\lambda_1}{2}}{\|\textbf{\emph{F}}-\textbf{\emph{Y}}\textbf{\emph{C}}\|}_F^2+{\| \textbf{\emph{C}}\|}_*,
\end{equation}
where $\textbf{\emph{C}}\in{\mathbb{R}}^{N\times N}$ is the coefficient matrix of a representation of $\textbf{\emph{Y}}\in{\mathbb{R}}^{n\times N}$ over itself. Typically, $\textbf{\emph{Y}}$ is chosen as the $\textbf{\emph{F}}$ itself. $\lambda_1$ is a tunable parameters which can be set according to the properties of norms. ${\|\textbf{\emph{C}}\|}_*$ is the nuclear norm of matrix. $\| \cdot \|_F^2$ represents the squared Frobenius norm.  %and $\| \cdot \|_1$ is the $\ell_1$-norm.

However, solving LRR is a challenge in terms of time complexity and memory cost. To remedy this issue, one potential way to solve this problem is an online manner namely noise-free online LRR (NOLRR). The matrix ${\|\textbf{\emph{C}}\|}_*$ can be denoted:
\begin{equation}
\|\textbf{\emph{C}}\|_*=\mathop{\min}_{\textbf{\emph{U}},\textbf{\emph{V}},\textbf{\emph{C}}=\textbf{\emph{U}}\textbf{\emph{V}}^\mathsf{T}}\dfrac{1}{2}
\Big(\|\textbf{\emph{U}}\|_F^2+\|\textbf{\emph{V}}\|_F^2\Big),
\end{equation}
where $\textbf{\emph{U}}\in{\mathbb{R}}^{N\times d}$ and $\textbf{\emph{V}}\in{\mathbb{R}}^{N\times d}$. To decouple the rows of $\textbf{\emph{U}}$, we use an auxiliary matrix $\textbf{\emph{D}}=\textbf{\emph{Y}}\textbf{\emph{U}}$ and approximate the term $\textbf{\emph{F}}$ with $\textbf{\emph{D}}\textbf{\emph{V}}^\mathsf{T}$. The matrix $\textbf{\emph{D}}$ is regarded as a basis dictionary of the data, $\textbf{\emph{V}}$ with being the coefficients. So we obtain a \emph{regularized} version of LRR and solve the following problem:
\begin{equation}
\label{eq:rverolrr}
\begin{split}
\mathop{\min}_{\textbf{\emph{D,U,V}}}{\dfrac{\lambda_1}{2}}{\|\textbf{\emph{F}}-\textbf{\emph{D}}\textbf{\emph{V}}^\mathsf{T}\|}_F^2+\dfrac{1}{2}
\Big(\|\textbf{\emph{U}}\|_F^2&+\|\textbf{\emph{V}}\|_F^2\Big)\\ &+{\dfrac{\lambda_2}{2}}\|\textbf{\emph{D}}-\textbf{\emph{Y}}\textbf{\emph{U}}\|_F^2.
\end{split}
\end{equation}

Let $\textbf{\emph{f}}_i$, $\textbf{\emph{y}}_i$, $\textbf{\emph{u}}_i$, and $\textbf{\emph{v}}_i$, be the $i$-th column of matrices $\textbf{\emph{F}}$, $\textbf{\emph{Y}}$, $\textbf{\emph{U}}^\mathsf{T}$, and $\textbf{\emph{V}}^\mathsf{T}$, respectively. Solving Eq.~(\ref{eq:rverolrr}) indeed minimizes the following empirical cost function,
\begin{equation}
\label{eq:emcostfun}
f_N(\textbf{\emph{D}})\triangleq\frac{1}{N}\sum\limits_{i=1}^N{\ell_1}(\textbf{\emph{f}}_i,\textbf{\emph{D}})+\frac{1}{N}\sum\limits_{i=1}^N{\ell_2}(\textbf{\emph{y}}_i,\textbf{\emph{D}}),
\end{equation}
where the loss function ${\ell_1}$, ${\ell_2}$ are defined as
\begin{equation}
\label{eq:l1}
{\ell_1}(\textbf{\emph{f}}_i,\textbf{\emph{D}})=\mathop{\min}_{\textbf{\emph{v}}}{\hat{\ell_1}}(\textbf{\emph{f}}_i,\textbf{\emph{D}},\textbf{\emph{v}}),
\end{equation}
\begin{equation}
\label{eq:l2}
{\ell_2}(\textbf{\emph{y}}_i,\textbf{\emph{D}})=\mathop{\min}_{\textbf{\emph{u}}}{\hat{\ell_2}}(\textbf{\emph{y}}_i,\textbf{\emph{D}},\textbf{\emph{M}}_{i-1},\textbf{\emph{u}}),
\end{equation}
that is
\begin{equation}
{\hat{\ell_1}}(\textbf{\emph{f}}_i,\textbf{\emph{D}},\textbf{\emph{v}})
\triangleq \frac{\lambda_1}{2}{\|\textbf{\emph{f}}_i-\textbf{\emph{D}}\textbf{\emph{v}}\|}_2^2+\frac{1}{2}{\|\textbf{\emph{v}}\|}_2^2,
\end{equation}
\begin{equation}
{\hat{\ell_2}}(\textbf{\emph{y}}_i,\textbf{\emph{D}},\textbf{\emph{M}}_{i-1},\textbf{\emph{u}})
\! \triangleq \! \frac{1}{2}{\|\textbf{\emph{u}}\|}_2^2+\frac{\lambda_2}{2}{\|\textbf{\emph{D}}-\textbf{\emph{M}}_{i-1}-\textbf{\emph{y}}_i\textbf{\emph{u}}^\mathsf{T}\|}_F^2,
\end{equation}
where
\begin{equation}
\textbf{\emph{M}}_{i-1}=\sum\limits_{j=1}^{i-1}\textbf{\emph{y}}_j\textbf{\emph{u}}_j^\mathsf{T}.
\end{equation}

In the $t$-th time instance, the objective function for updating the basis $\textbf{\emph{D}}$ is defined as
\begin{equation}
g_t(\textbf{\emph{D}})
\! \triangleq \! \frac{1}{t}\Big(\!\sum\limits_{i=1}^t \! {\hat{\ell_1}}(\textbf{\emph{f}}_i,\textbf{\emph{D}},\textbf{\emph{v}}_i)+\! \sum\limits_{i=1}^t \! \frac{1}{2}{\|\textbf{\emph{u}}_i\|}_2^2+\frac{\lambda_2}{2}{\|\textbf{\emph{D}}-\textbf{\emph{M}}_t\|}_F^2\Big).
\end{equation}
This is a surrogate function of the empirical cost function $f_t(\textbf{\emph{D}})$ defined in Eq.~(\ref{eq:emcostfun}), namely it provides an upper bound for $f_t(\textbf{\emph{D}}): g_t(\textbf{\emph{D}}) \geqslant f_t(\textbf{\emph{D}})$. Expanding the first term, $\textbf{\emph{D}}_t$ is given by:
\begin{equation}
\label{eq:dt}
\begin{split}
\textbf{\emph{D}}_t=\mathop{\arg\min}_{\textbf{\emph{D}}}{\dfrac{1}{t}}\Big[\dfrac{1}{2}\mathop{\mathrm{Tr}}&(\textbf{\emph{D}}^\mathsf{T}\textbf{\emph{D}}(\lambda_1\textbf{\emph{A}}_t
+\lambda_2\textbf{\emph{I}}_d) \\ & -\mathop{\mathrm{Tr}}(\textbf{\emph{D}}^\mathsf{T}(\lambda_1\textbf{\emph{B}}_t+\lambda_2\textbf{\emph{M}}_t))\Big],
\end{split}
\end{equation}
where $\textbf{\emph{A}}_t=\sum_{i=1}^t\textbf{\emph{v}}_i\textbf{\emph{v}}_i^\mathsf{T}$ and $\textbf{\emph{B}}_t=\sum_{i=1}^t\textbf{\emph{f}}_i\textbf{\emph{v}}_i^\mathsf{T}$. In practice, a block coordinate descent approach~\cite{shen2016online} can be applied to minimize over $\textbf{\emph{D}}$.

Usually, a potential way to construct a graph is collecting all the $\textbf{\emph{u}}_i$ and $\textbf{\emph{v}}_i$ to compute the representation matrix $\textbf{\emph{C}}=\textbf{\emph{U}}\textbf{\emph{V}}^\mathsf{T}$. So the NOLRR-graph is built as $
\emph{\textbf{W}}=(|\textbf{\emph{C}}|+|{\textbf{\emph{C}}}|^\mathsf{T})/2$.
The above NOLRR reduces the memory cost from \emph{O}($N^2$) to \emph{O}($nd$) with $d$ being estimated rank ($d<n\ll N$) and makes it an appealing solution for large scale data. The time complexity of the update on accumulation matrices is \emph{O}($nd$) and that of $\textbf{\emph{D}}_t$ is \emph{O}($nd^2$). The procedure of NOLRR-graph construction is summarized in \textbf{Algorithm~\ref{alg:olrrgraph}}. The NOLRR-graph can improve the performance of natural image segmentation as explained in Section~\ref{subsec:different modules}.

\renewcommand{\algorithmicrequire}{\textbf{Input:}}
\renewcommand{\algorithmicensure}{\textbf{Output:}}
\begin{algorithm}[!t]
	\caption{Global node selection}
	\label{alg:global nodes selection}
	\begin{algorithmic}[1]
		\REQUIRE Feature $\textbf{\emph{F}}\in \mathbb{R}^{n\times N}$, $e$, $g$, $\psi=3$, $\tau=10^{-6}$ ;
		\STATE Compute $\textbf{\emph{c}}_j^{*}$ from OMP$(\textbf{\emph{F}}_{-j},\textbf{\emph{f}}_j)$;
		\STATE Set $\textbf{\emph{C}}^{*}=[\textbf{\emph{c}}_1^{*},...,\textbf{\emph{c}}_N^{*}]$ and $\textbf{\emph{M}}_{sp}=|\textbf{\emph{C}}^{*}|+|{\textbf{\emph{C}}^{*}}^\mathsf{T}|$;
		\STATE Compute classification from  $\textbf{\emph{M}}_{sp}$ by spectral clustering with the clustering number provided by the APC.
		\ENSURE Global nodes.
	\end{algorithmic}
\end{algorithm}

\renewcommand{\algorithmicrequire}{\textbf{Input:}}
\renewcommand{\algorithmicensure}{\textbf{Output:}}
\begin{algorithm}[!t]
\caption{NOLRR-graph construction}
\label{alg:olrrgraph}
\begin{algorithmic}[1]
\REQUIRE Feature $\textbf{\emph{F}}\in \mathbb{R}^{n\times N}$; parameter $d$;
\STATE \textbf{Initialize} $\lambda_1=1$, $\lambda_2^{ini}=1/\sqrt{n}$, basis dictionary $\textbf{\emph{D}}_0\in \mathbb{R}^{n\times d}$, zero matrices $\textbf{\emph{A}}_0\in \mathbb{R}^{d\times d}$, $\textbf{\emph{B}}_0\in \mathbb{R}^{n\times d}$, $\textbf{\emph{M}}_0\in \mathbb{R}^{n\times d}$, $\textbf{\emph{U}}\in \mathbb{R}^{N\times d}$, $\textbf{\emph{V}}\in \mathbb{R}^{N\times d}$.
\FOR {$t=1$ to $N$}
    \STATE  Access the $t$-th atom $\textbf{\emph{f}}_t$ and compute $\lambda_2=\sqrt{t}\times \lambda_2^{ini}$;
    \STATE  Compute the coefficients $\textbf{\emph{v}}_t$ and $\textbf{\emph{u}}_t$ using Eq.~(\ref{eq:l1}) and Eq.~(\ref{eq:l2}), respectively;
    %\begin{center}
    % $\textbf{\emph{v}}_t=\mathop{\arg\min}_{\textbf{\emph{v}}}{\hat{\ell_1}}(\textbf{\emph{f}}_t,\textbf{\emph{D}}_{t-1},\textbf{\emph{v}})$,
    % $\textbf{\emph{u}}_t=\mathop{\arg\min}_{\textbf{\emph{u}}}{\hat{\ell_2}}(\textbf{\emph{f}}_t,\textbf{\emph{D}}_{t-1},\textbf{\emph{M}}_{t-1},\textbf{\emph{u}})$;
    %\end{center}
    \STATE  Update the accumulation matrices
    \begin{center}
     $\textbf{\emph{A}}_t \leftarrow \textbf{\emph{A}}_{t-1}+\textbf{\emph{v}}_{t}\textbf{\emph{v}}_{t}^\mathsf{T}$,
     $\textbf{\emph{B}}_t \leftarrow \textbf{\emph{B}}_{t-1}+\textbf{\emph{f}}_{t}\textbf{\emph{v}}_{t}^\mathsf{T}$,
     $\textbf{\emph{M}}_t \leftarrow \textbf{\emph{M}}_{t-1}+\textbf{\emph{f}}_{t}\textbf{\emph{v}}_{t}^\mathsf{T}$;
    \end{center}
    \STATE Update the basis dictionary using Eq.~(\ref{eq:dt});
    \STATE Update the matrices
    \begin{center}
     $\textbf{\emph{U}}_{t,:} \leftarrow \textbf{\emph{u}}_{t}$, $\textbf{\emph{V}}_{t,:} \leftarrow \textbf{\emph{v}}_{t}$;
    \end{center}
\ENDFOR
\STATE Compute the matrix $\textbf{\emph{C}}=\textbf{\emph{U}}\textbf{\emph{V}}^\mathsf{T}$ and the NOLRR-graph $\emph{\textbf{W}}=(|\textbf{\emph{C}}|+|{\textbf{\emph{C}}}|^\mathsf{T})/2$.
\ENSURE NOLRR-graph $\textbf{\emph{W}}$.
\end{algorithmic}
\end{algorithm}

%Based on superpixels, the proposed adaptive fusion affinity graph, and the bipartite graph, the overall AFA-graph scheme for natural image segmentation is summarized in \textbf{Algorithm~\ref{alg:afagraph}}.

%\renewcommand{\algorithmicrequire}{\textbf{Input:}}
%\renewcommand{\algorithmicensure}{\textbf{Output:}}
%\begin{algorithm} [!t]
%    \caption{AFA-graph for natural image segmentation}
%    \label{alg:afagraph}
%    \begin{algorithmic}[1]
%        \REQUIRE Source image $\textbf{\emph{I}}$, rank $d$, group $k_T$.
%        \STATE Over-segment the image $I$ by the SAS and obtain superpixels at different scales;
%        \STATE Construct an adjacency-graph by the mLab features extracted from all superpixels at each scale;
%        \STATE Select global nodes of superpixels through APC results based on the SPR;
%        \STATE Build an OLRR-graph by the mLab features extracted from the global nodes at each scale;
%        \STATE Update the adjacency-graph by the OLRR-graph at each scale;
%        \STATE Fuse the updated graphs across different scales to obtain the final segmentation result (pixel-wise labels) through Tcut with the group $k_T$.
%        \ENSURE Pixel-wise labels.
%        \end{algorithmic}
%\end{algorithm}

%% file: files/EXPERIMENTS.tex
\section{EXPERIMENTS AND ANALYSIS}
To verify the performance of the proposed AFA-graph, we first introduce the experimental setup, and then show the results of our approach and its variations. Moreover, the results of our method are presented compared with the state-of-the-art methods on different datasets. Finally, we show time complexity analysis of our AFA-graph.

\subsection{Experimental Setup}
\subsubsection{Databases}
All experiments are carried out on the following publicly available databases: Berkeley Segmentation Database (BSD)~\cite{martin2001database} and Microsoft Research Cambridge (MSRC) database~\cite{Shotton2006TextonBoost}, Stanford Background Dataset (SBD)~\cite{5459211}, and PASCAL visual object classes 2012 (VOC) dataset~\cite{Everingham10}.
\begin{itemize}
  \item BSD300: The BSD300 includes 300 natural images and the ground truth data (each one has about 5 human annotations). Each image has a fixed size of 481$\times$321 pixels.
  \item BSD500: As an improved version of BSD300 dataset, the BSD500 contains 500 natural images. Each image is annotated by 5 different people on average.
  \item MSRC: The MSRC contains 591 images and 23 object classes with accurate pixel-wise labeled images. The performance is evaluated using the clean ground-truth object instance labeling of~\cite{Malisiewicz2007Improving}.
  \item SBD: The SBD contains 715 images of urban and rural scenes with 8 classes. Each image is approximately 240$\times$320 pixels and contains at least one foreground object. The `layers', `regions', and `surfaces' are all used as groundtruth.
  \item PASCAL VOC: The segmentation challenge of PASCAL VOC12 includes 2913 images (1416 images in train set and 1449 images in val set) with annotated objects of 20 categories. Following the standard setting, we use val set to test the final performance.
\end{itemize}

\subsubsection{Metrics}
To evaluate our method, there are four standard measurements: the probabilistic rand index (PRI)~\cite{unnikrishnan2007toward}, the variation of information (VoI)~\cite{meilua2007comparing}, the global consistency error (GCE)~\cite{martin2001database}, and the boundary displacement error (BDE)~\cite{freixenet2002yet}. 
The closer the segmentation result is to the ground truth, the higher the PRI is, and the smaller the other three measures (VoI, GCE, and BDE) are. All experiments are conducted under the PC condition of 3.40GHz of Intel Xeon E5-2643 v4 processor, 64G RAM, Ubuntu 16.04 OS and Matlab 2018a.

\subsection{Performance analysis for framework}
\label{subsec:different modules}
\subsubsection{Influence of parameter $\alpha$ in IKDE}
\begin{figure}[!t]
	\centering
	\subfloat[]{
		\includegraphics[width=1.65in]{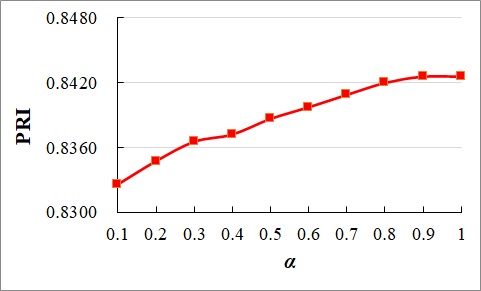}
	}
	\subfloat[]{
		\includegraphics[width=1.65in]{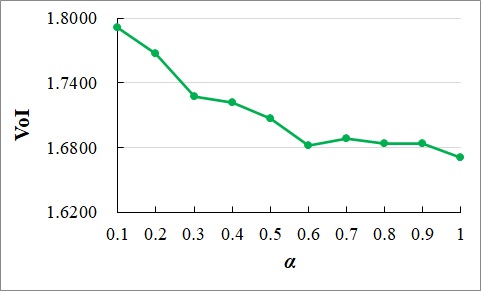}
	}
	
	\vspace{-0.5em}
	\subfloat[]{
		\includegraphics[width=1.65in]{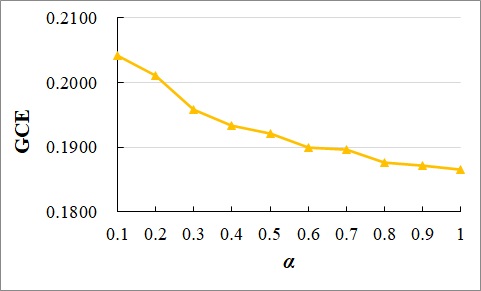}
	}
	\subfloat[]{
		\includegraphics[width=1.65in]{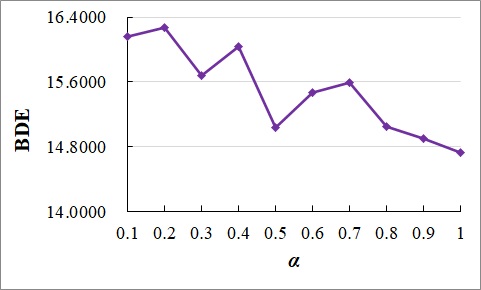}
	}
	\caption{Examine the influence of $\alpha$ in IKDE on BSD300 dataset. To accurately describe the influence of $\alpha$ on performance, the values of four indexes retain four decimal places. (a) PRI. (b) VoI. (c) GCE. (d) BDE.}
	\label{fig:alphainfluence}
\end{figure}

To analyze the robustness of our AFA-graph, we first explore the influence of exponential factor $\alpha$ in IKDE. As shown in Figs.~\ref{fig:alphainfluence}(a)-(d), we report the indexes of PRI, VoI, GCE, and BDE for different $\alpha$ under a range of corruptions on BSD300 dataset. From Fig.~\ref{fig:alphainfluence}, we can find that the best performance is obtained when parameter $\alpha=1$. The index of PRI is highest and other three indexes are closed to the best results.

\subsubsection{Influence of parameters $e$ and $g$ in APC}
\begin{figure}[!t]
	\centering
	\subfloat[]{
		\includegraphics[width=1.65in]{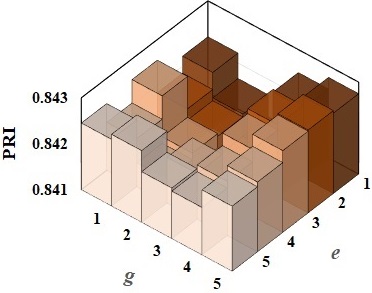}
	}
	\subfloat[]{
		\includegraphics[width=1.65in]{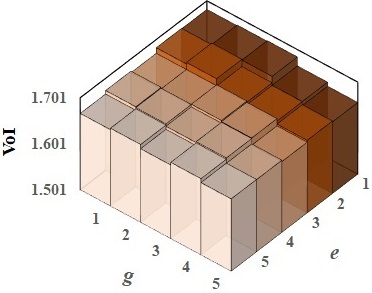}
	}
	
	\vspace{-0.5em}
	\subfloat[]{
		\includegraphics[width=1.65in]{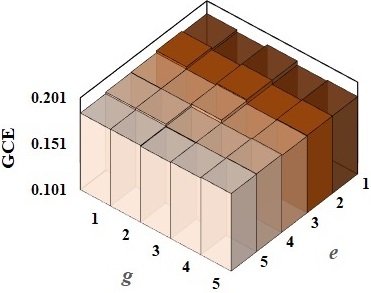}
	}
	\subfloat[]{
		\includegraphics[width=1.65in]{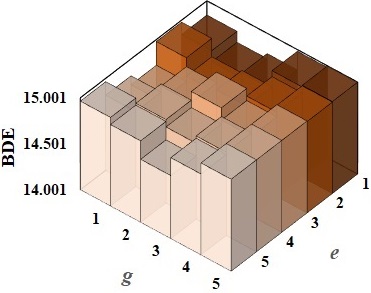}
	}
	\caption{Examine the influence of $e$ and $g$ in APC on BSD300 dataset. (a) PRI. (b) VoI. (c) GCE. (d) BDE.  Our method has good robustness to the variety of parameter.}
	\label{fig:APCinfluence}
\end{figure}

To analyze the robustness of our AFA-graph, we then explore the influence of parameters $e$ and $g$ in APC. We report the results for different $e$ and $g$ under a range of corruptions on BSD300 dataset as shown in Figs.~\ref{fig:APCinfluence}(a)-(d). From the results, we can find that the best performance is obtained when parameter $e=3$ and $g=5$. Our method has good robustness to the variety of parameters $e$ and $g$.

\subsubsection{Influence of parameter $d$ in NOLRR}
\begin{figure}[!t]
	\centering
	\subfloat[]{
		\includegraphics[width=1.65in]{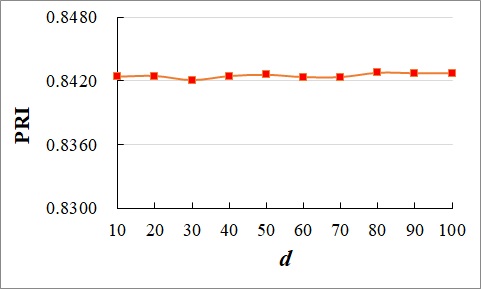}
	}
	\subfloat[]{
		\includegraphics[width=1.65in]{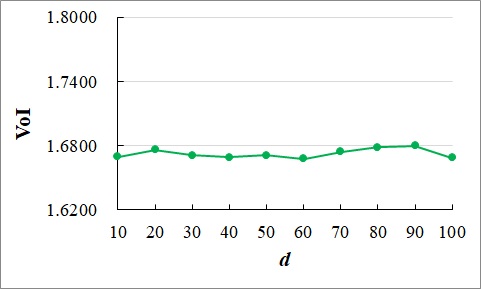}
	}
	
	\vspace{-0.5em}
	\subfloat[]{
		\includegraphics[width=1.65in]{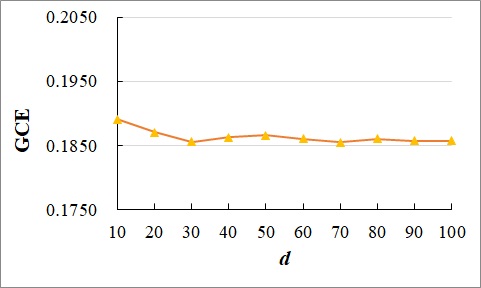}
	}
	\subfloat[]{
		\includegraphics[width=1.65in]{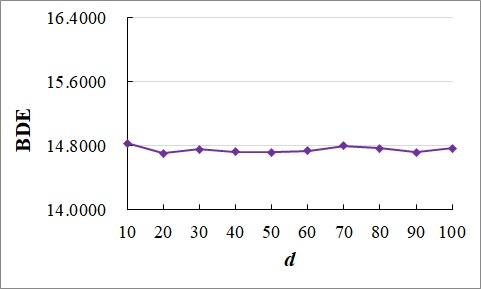}
	}
	\caption{Examine the influence of $d$ in NOLRR on BSD300 dataset. To accurately describe the influence of $d$ on performance, the values of four indexes retain four decimal places. (a) PRI. (b) VoI. (c) GCE. (d) BDE. Our method has good robustness to the variety of parameter.}
	\label{fig:dinfluence}
\end{figure}

To analyze the robustness of our AFA-graph, we also explore the influence of the selected rank $d$ in the proposed NOLRR (\textbf{Algorithm~\ref{alg:olrrgraph}}). In Figs.~\ref{fig:dinfluence}(a)-(d), our proposed NOLRR also has good robustness to the variety of parameter. The index of PRI is highest and other three indexes are closed to the best results. So, the best performance is obtained when parameter $d=50$.

\subsubsection{Denoise methods}
\begin{figure}[!t]
	\centering
	\includegraphics[width=0.84in]{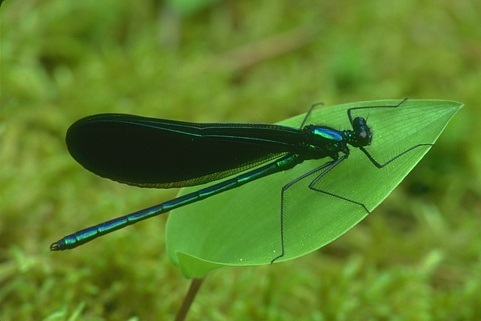}\hspace{-0.1em}
	\includegraphics[width=0.84in]{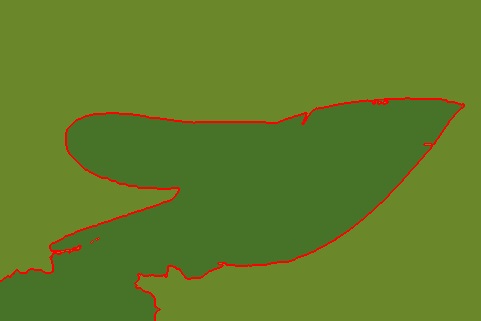}\hspace{-0.1em}
	\includegraphics[width=0.84in]{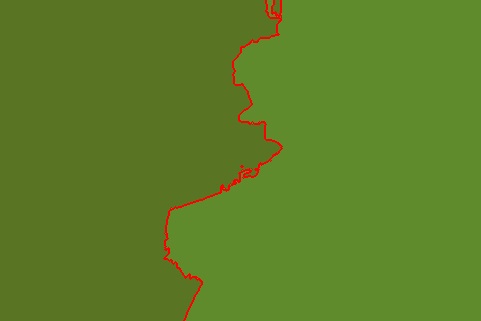}\hspace{-0.1em}
	\includegraphics[width=0.84in]{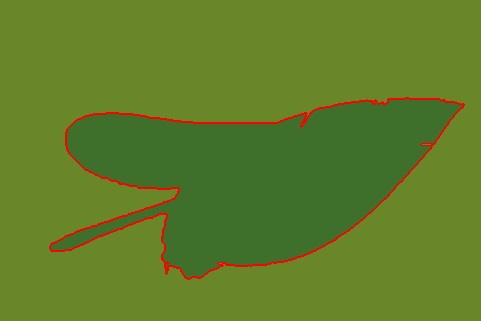}
	
	\vspace{0.2em}
	\includegraphics[width=0.84in]{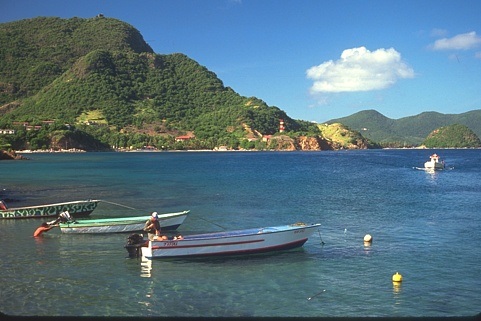}\hspace{-0.1em}
	\includegraphics[width=0.84in]{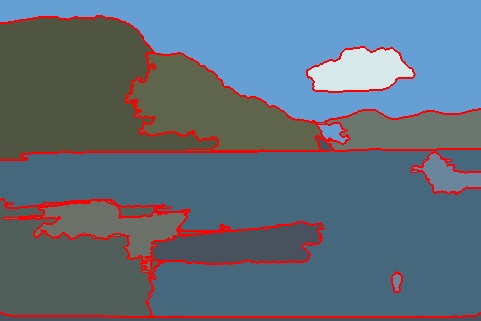}\hspace{-0.1em}
	\includegraphics[width=0.84in]{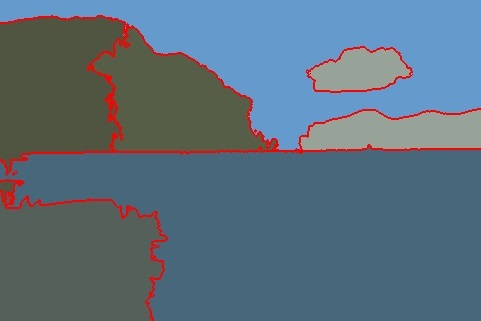}\hspace{-0.1em}
	\includegraphics[width=0.84in]{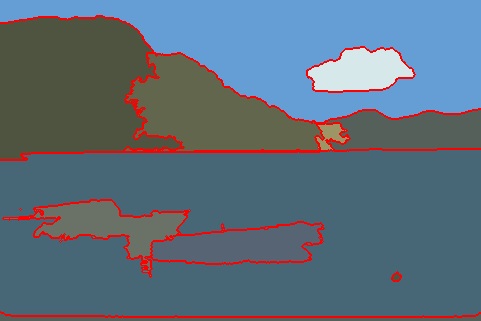}
		
	\vspace{0.2em}
	\includegraphics[width=0.84in]{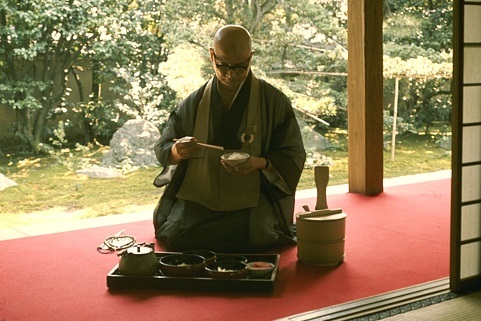}\hspace{-0.1em}
	\includegraphics[width=0.84in]{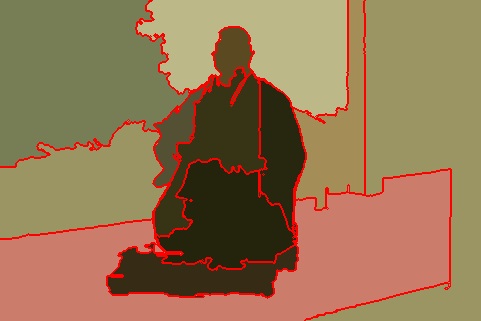}\hspace{-0.1em}
	\includegraphics[width=0.84in]{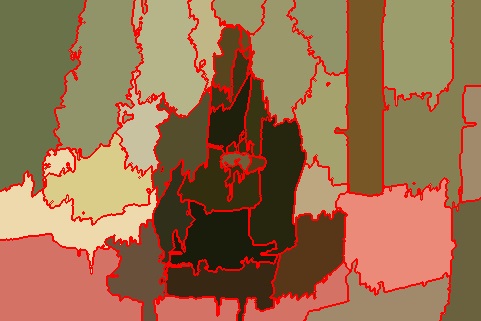}\hspace{-0.1em}
	\includegraphics[width=0.84in]{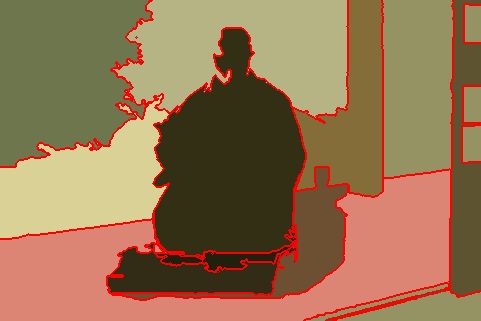}
		
	\vspace{0.2em}
	\includegraphics[width=0.84in]{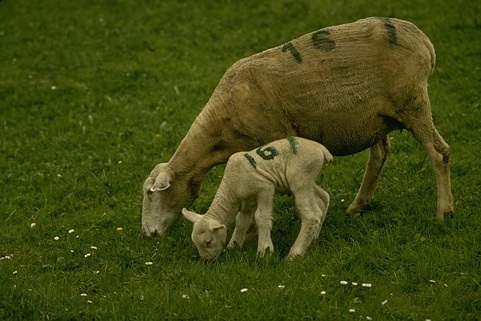}\hspace{-0.1em}
	\includegraphics[width=0.84in]{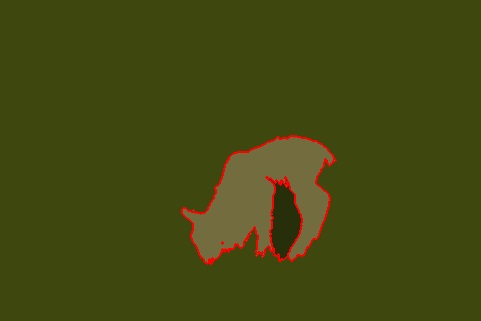}\hspace{-0.1em}
	\includegraphics[width=0.84in]{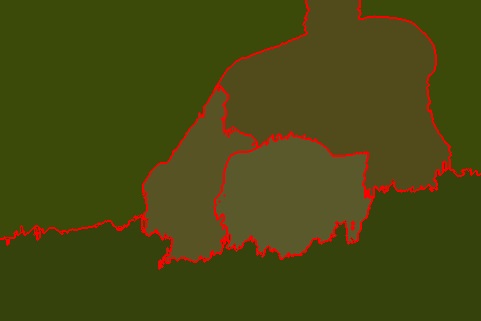}\hspace{-0.1em}
	\includegraphics[width=0.84in]{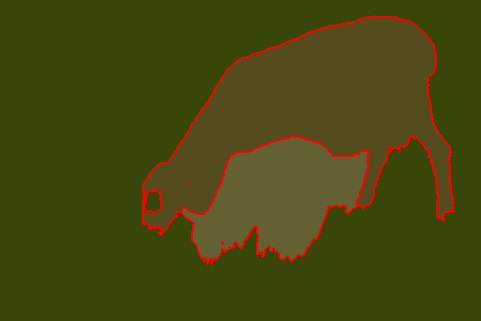}
	\caption{Visual comparison on BSD300 dataset obtained by our AFA-graph. From left to right, input images, the results of AFA-graph w/o IKDE, IKDE on images and IKDE on features are presented. The results of AFA-graph with IKDE on features are much better, in particular often more accurate.}
	\label{fig:graphs1}
\end{figure}

As suggested, we have conducted experiments to explore the influence of noise to accuracy. The results on BSD300 are listed in Table~\ref{different denoise}. The Gaussian means that we perform Gaussian filtering on images or features as done by SFFCM~\cite{Lei2018fuzzy}. The size of filter is [5, 5], and the value for sigma is set as 1. The Bilateral means that we perform shiftable bilateral filtering on images or features as done by CCP~\cite{7410546}. The size of filter is [5, 5], and the width of both spatial and range Gaussian are set as 5. For IKDE, the values of $\alpha$ is an exponential factor is set to 1. We can observe that IKDE can achieve the best results on all metrics. In Fig.~\ref{fig:graphs1}, the results also show that the performance of filtering on images is worse than filtering on features. In this case, filtering on images may influence the neighborhood topology between pixels which is the core of a constructed affinity graph.
\begin{table}[!t]
	\centering
	\renewcommand{\arraystretch}{1.3}
	\caption{Quantitative results of the proposed framework with different denoising methods on BSD300 dataset. We perform Gaussian, bilateral, and our IKDE filtering on images or features.}
	\label{different denoise}
	\begin{tabular}{lcccccc}
		\toprule
		Methods & Images & Features & PRI $\uparrow$  & VoI $\downarrow$ & GCE $\downarrow$ & BDE $\downarrow$ \\
		\midrule
		%None             &         &         & 0.83 & 1.68 & 0.18 & 14.71 \\
		Gaussian         & $\surd$ &         & 0.81 & 2.22 & 0.24 & 18.56 \\
		Gaussian         &         & $\surd$ & 0.84 & 1.76 & 0.20 & 15.71 \\
		Bilateral        & $\surd$ &         & 0.81 & 2.20 & 0.24 & 19.17 \\
		Bilateral        &         & $\surd$ & 0.84 & 1.68 & 0.18 & 15.26 \\ 
		IKDE             & $\surd$ &         & 0.81 & 2.22 & 0.25 & 19.17 \\
		IKDE             &         & $\surd$ & 0.84 & 1.67 & 0.19 & 14.72 \\
		\bottomrule
	\end{tabular}
\end{table}

\begin{table}[!t]
	\centering
	\renewcommand{\arraystretch}{1.3}
	\caption{Quantitative results of the proposed NOLRR-graph with different basic graphs using mLab features on BSD300 dataset. A-graph means adjacency-graph.}
	\label{different visual fea}
	\begin{tabular}{lccccc}
		\toprule
		Methods & IKDE & PRI $\uparrow$  & VoI $\downarrow$ & GCE $\downarrow$ & BDE $\downarrow$ \\
		\midrule
		A-graph                  &         & 0.83 & 1.75 & 0.18 & 15.02 \\
		A-graph                  & $\surd$ & 0.83 & 1.75 & 0.18 & 14.94 \\
		%LRR-graph        &         & 0.84 & 1.75 & 0.20 & 15.54 \\
		%LRR-graph        & $\surd$ & 0.84 & 1.74 & 0.20 & 15.47 \\
		%OLRSC-graph              &         & 0.81 & 2.32 & 0.25 & 17.59 \\
		%OLRSC-graph              & $\surd$ & 0.81 & 2.33 & 0.25 & 17.56 \\
		NOLRR-graph              &         & 0.81 & 2.32 & 0.25 & 17.57 \\
		NOLRR-graph              & $\surd$ & 0.81 & 2.34 & 0.24 & 17.42 \\
		%A- + LRR-graph   & $\surd$ & 0.84 & 1.66 & 0.18 & 14.84 \\
	    %A- + OLRSC-graph         &         & 0.84 & 1.68 & 0.19 & 14.74 \\
	    %A- + OLRSC-graph         & $\surd$ & 0.84 & 1.67 & 0.19 & 14.72 \\
	    A- + NOLRR-graph         &         & 0.84 & 1.68 & 0.19 & 14.74 \\
	    A- + NOLRR-graph         & $\surd$ & 0.84 & 1.67 & 0.19 & 14.72 \\	
		\bottomrule
	\end{tabular}
\end{table}

\subsubsection{Graphs}
\begin{figure*}[!t]
	\centering
	\includegraphics[height=0.77in]{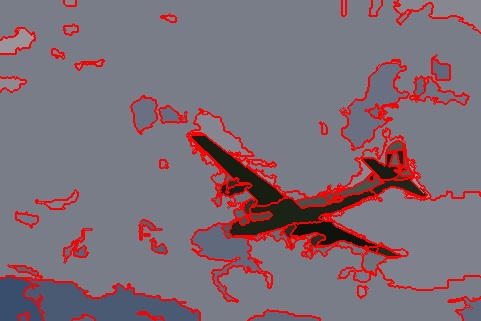}\hspace{-0.1em}
	\includegraphics[height=0.77in]{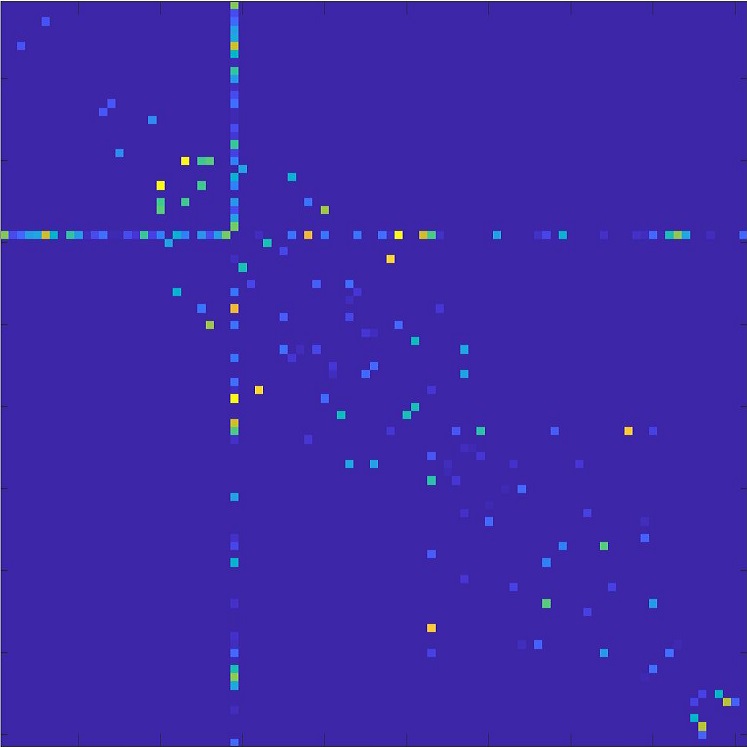}\hspace{-0.1em}
	\includegraphics[height=0.77in]{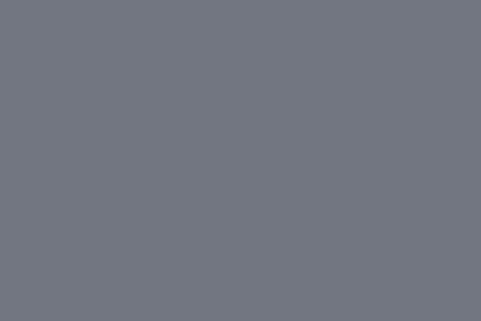}\hspace{-0.1em}
	\includegraphics[height=0.77in]{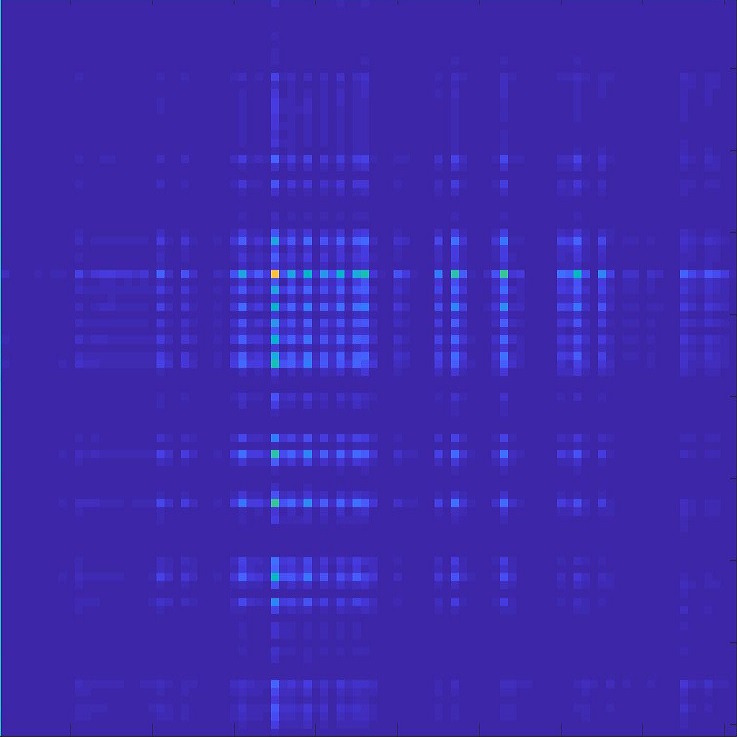}\hspace{-0.1em}
	\includegraphics[height=0.77in]{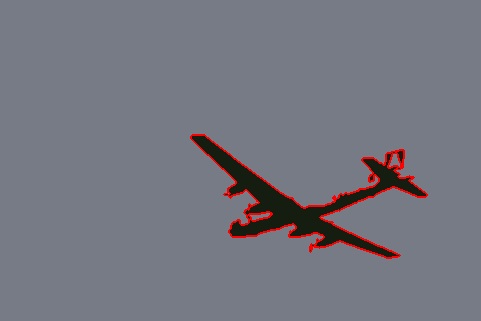}\hspace{-0.1em}
	\includegraphics[height=0.77in]{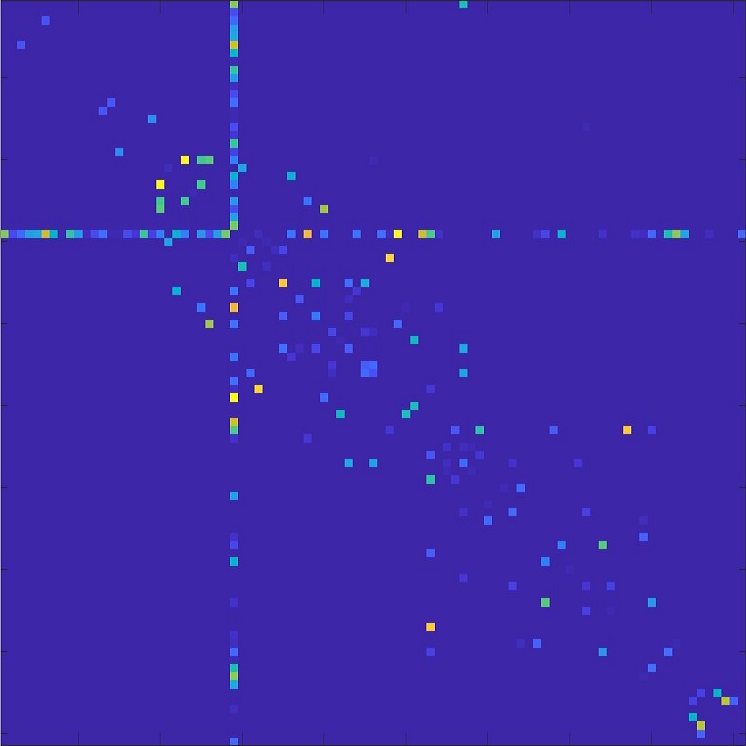}\hspace{-0.1em}
	\includegraphics[height=0.77in]{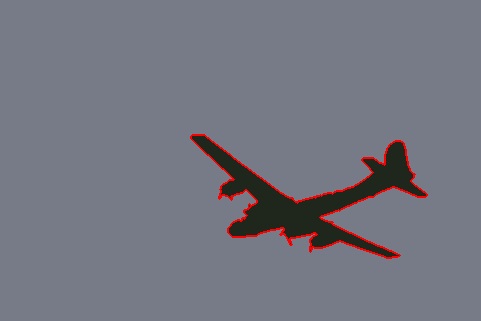}
	
	\vspace{0.3em}
	\includegraphics[height=0.77in]{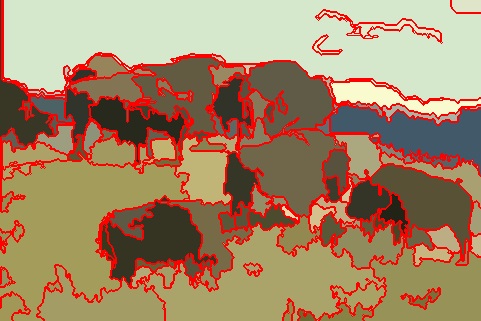}\hspace{-0.1em}
	\includegraphics[height=0.77in]{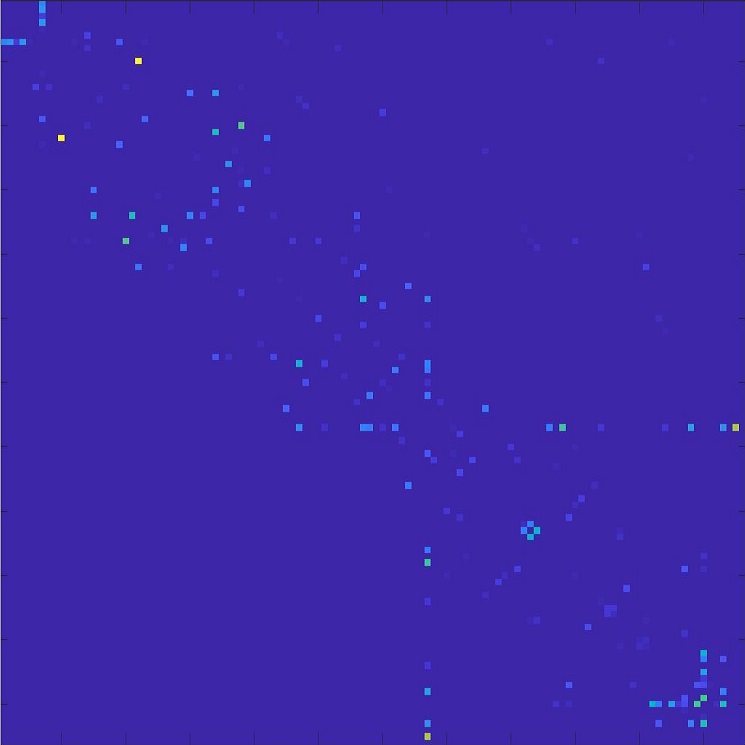}\hspace{-0.1em}
	\includegraphics[height=0.77in]{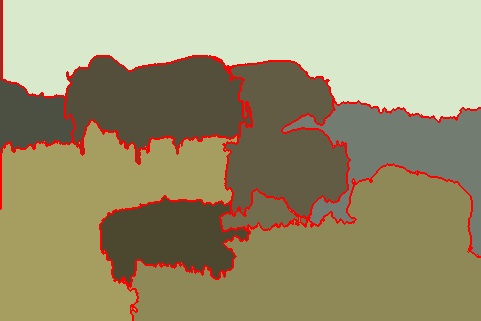}\hspace{-0.1em}
	\includegraphics[height=0.77in]{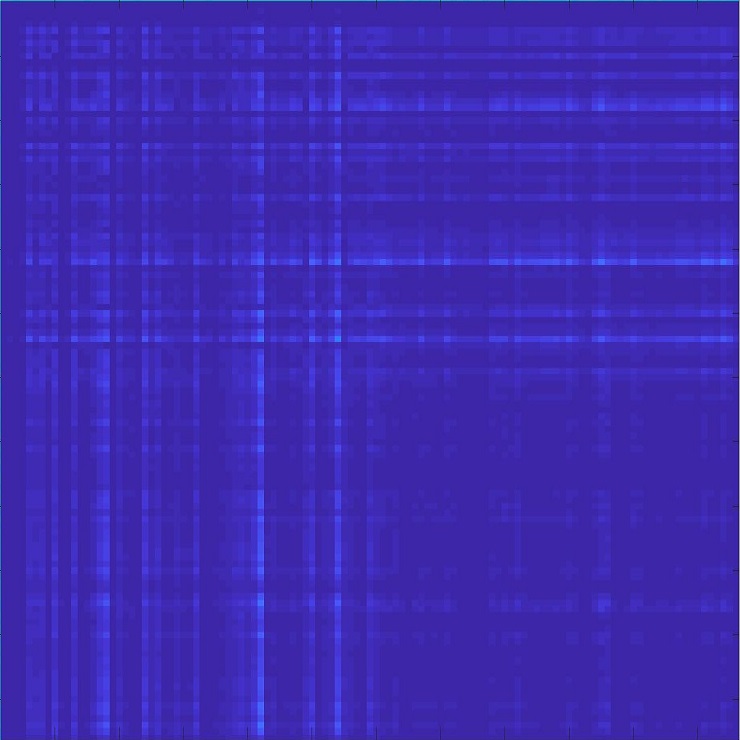}\hspace{-0.1em}
	\includegraphics[height=0.77in]{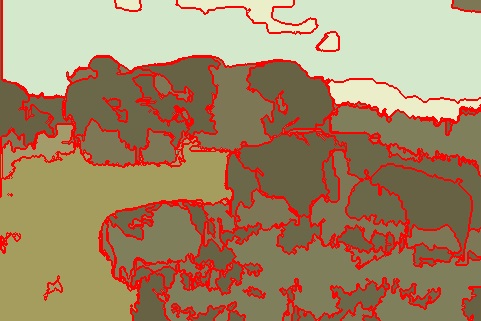}\hspace{-0.1em}
	\includegraphics[height=0.77in]{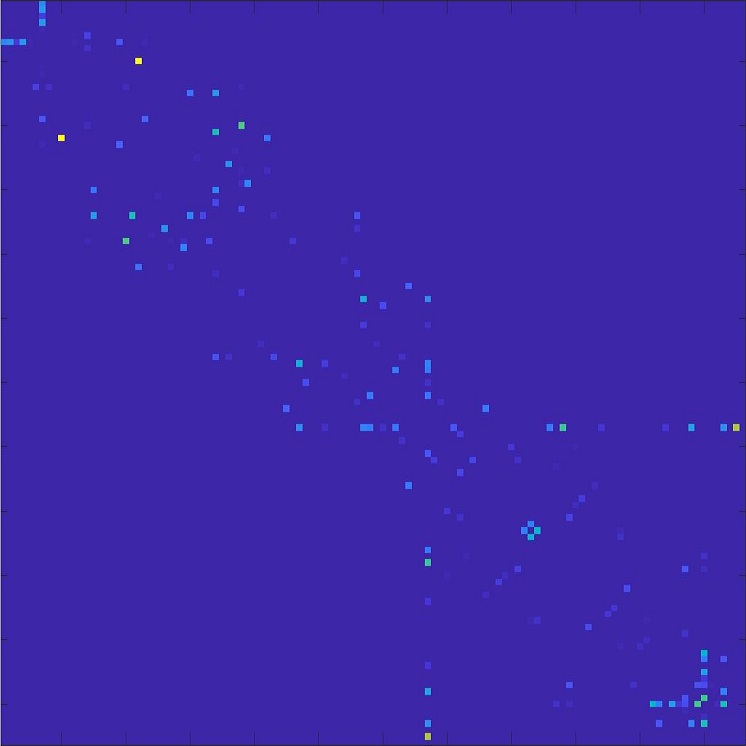}\hspace{-0.1em}
	\includegraphics[height=0.77in]{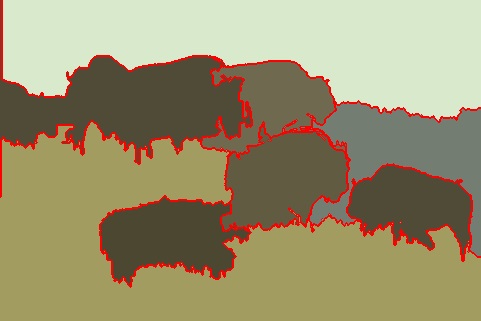}
	
	\vspace{0.3em}
	\includegraphics[height=0.77in]{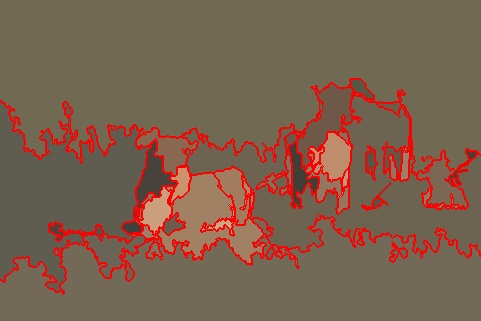}\hspace{-0.1em}
	\includegraphics[height=0.77in]{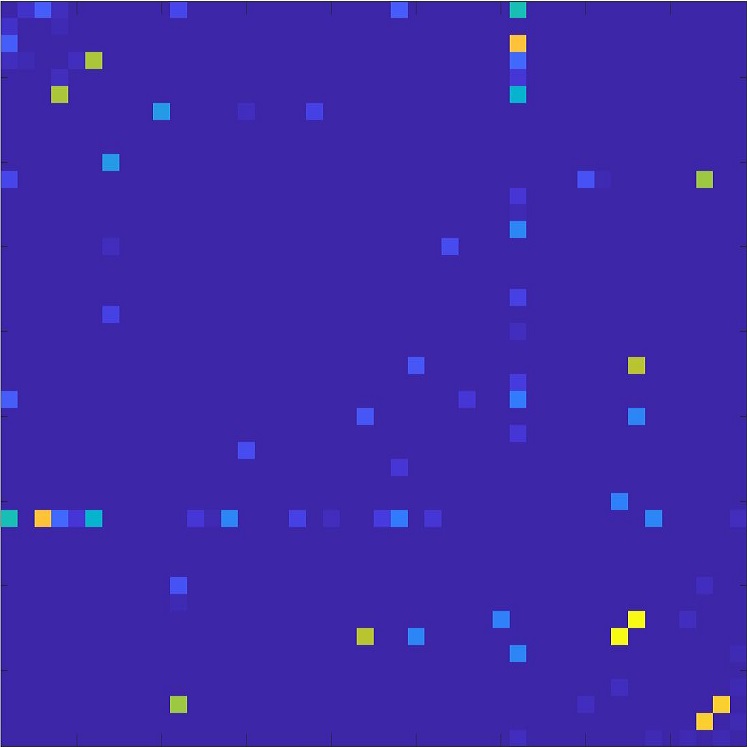}\hspace{-0.1em}
	\includegraphics[height=0.77in]{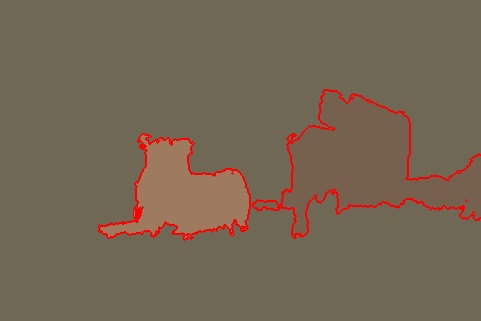}\hspace{-0.1em}
	\includegraphics[height=0.77in]{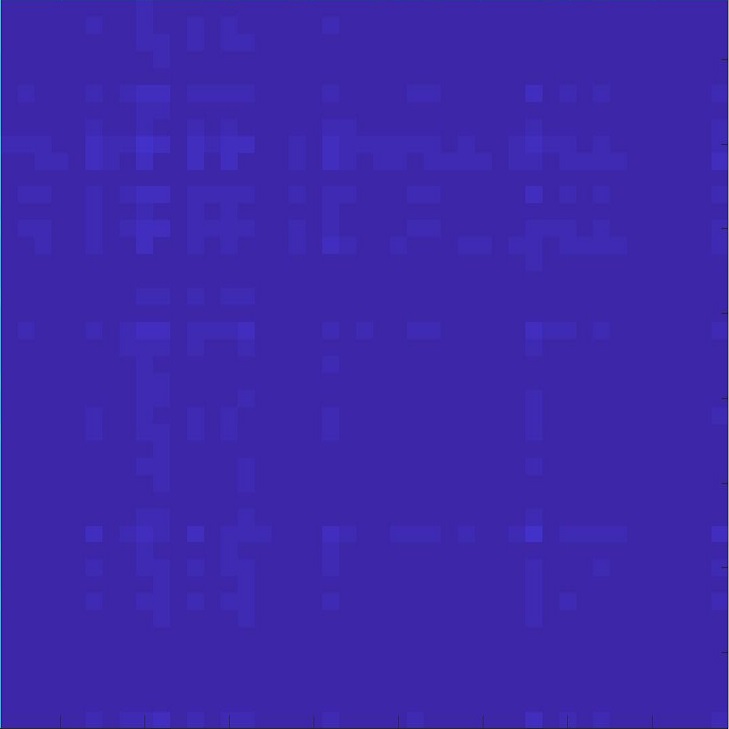}\hspace{-0.1em}
	\includegraphics[height=0.77in]{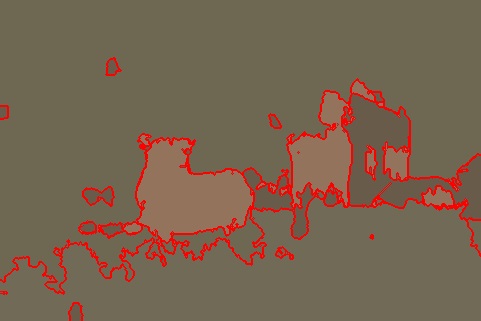}\hspace{-0.1em}
	\includegraphics[height=0.77in]{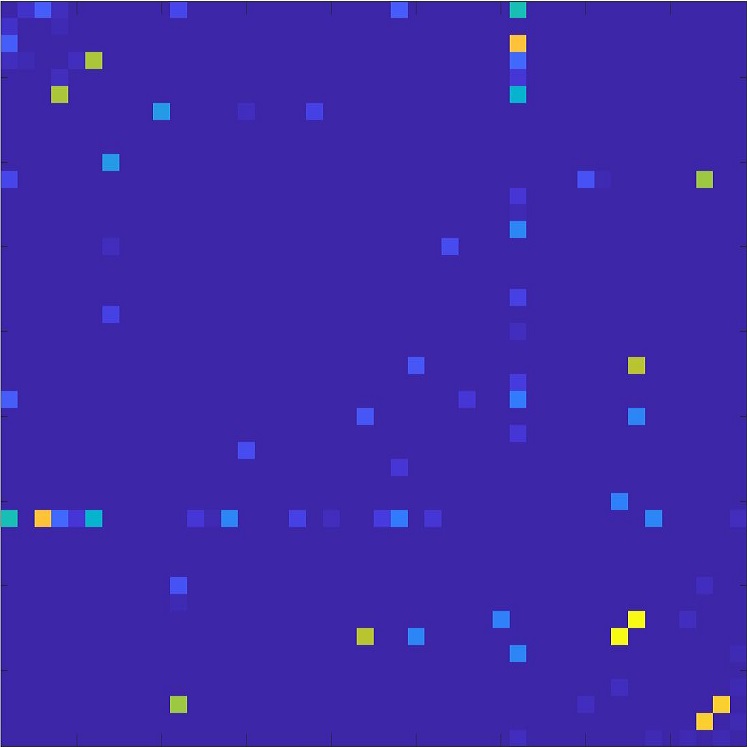}\hspace{-0.1em}
	\includegraphics[height=0.77in]{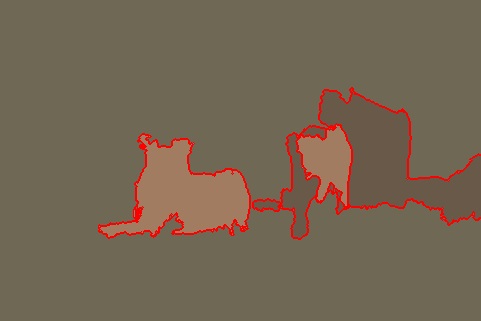}
	
	\vspace{0.3em}	
	\includegraphics[height=0.77in]{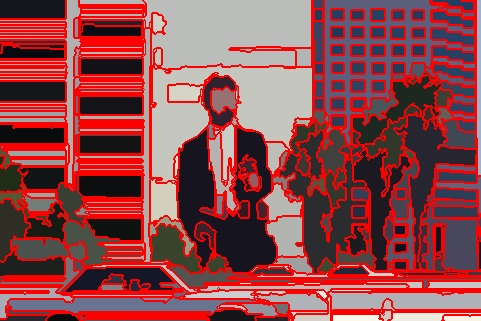}\hspace{-0.1em}
	\includegraphics[height=0.77in]{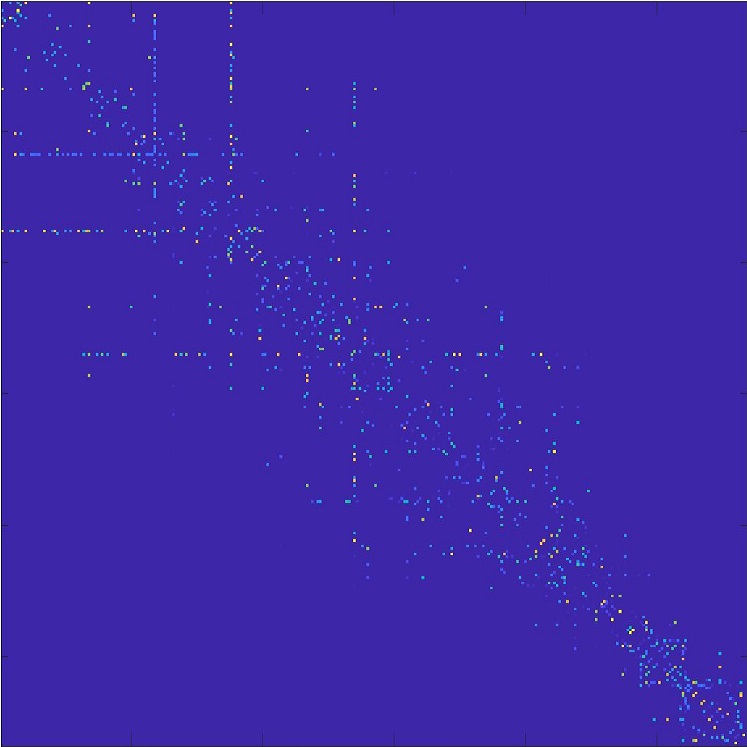}\hspace{-0.1em}
	\includegraphics[height=0.77in]{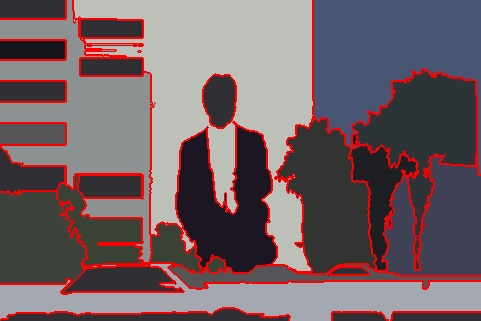}\hspace{-0.1em}
	\includegraphics[height=0.77in]{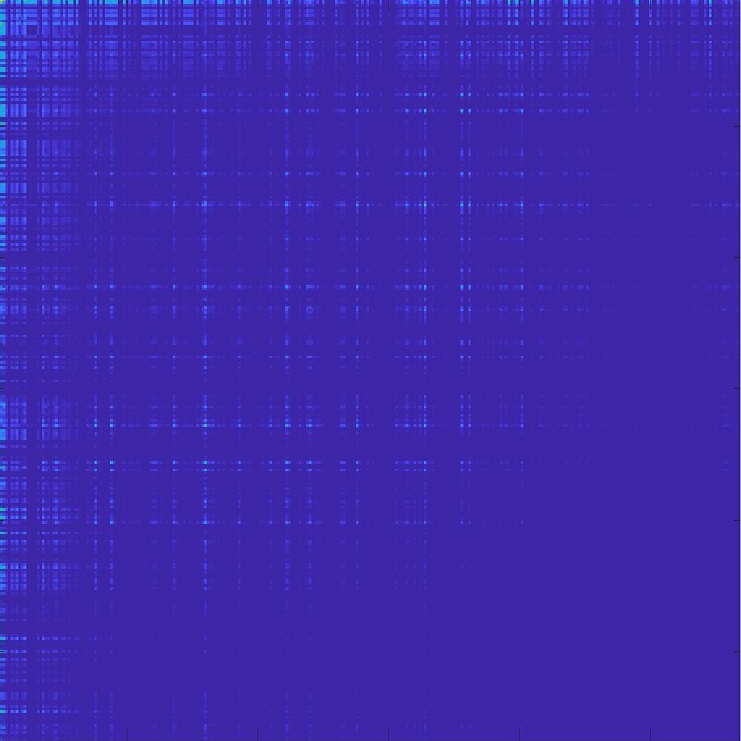}\hspace{-0.1em}
	\includegraphics[height=0.77in]{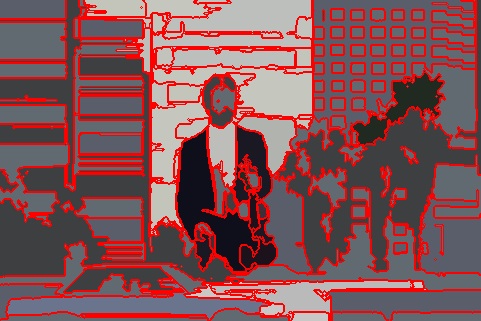}\hspace{-0.1em}
	\includegraphics[height=0.77in]{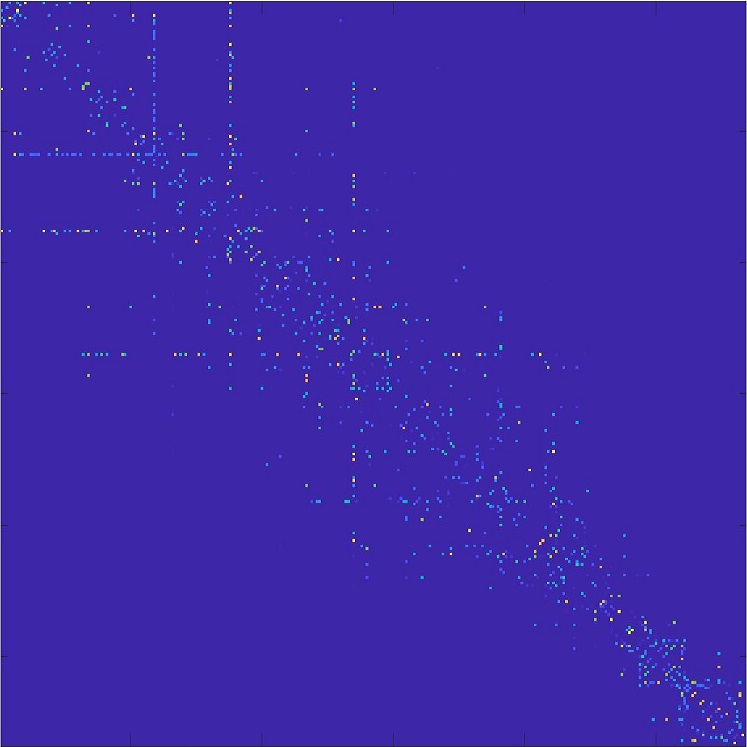}\hspace{-0.1em}
	\includegraphics[height=0.77in]{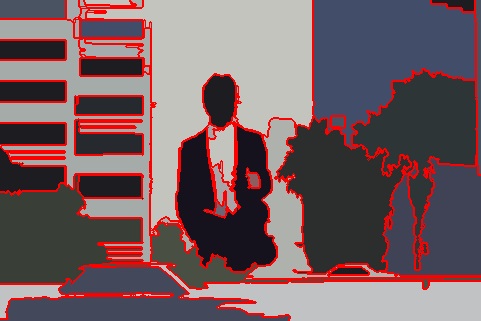}
	\caption{Visual comparison on BSD300 dataset obtained by different affinity graphs. From left to right, input superpixels, A-graph with the segmentation results, NOLRR-graph with the  results, and our AFA-graph (A- + NOLRR-graph) with the results are presented. Our AFA-graph combines different affinity graphs with assimilating the advantages of two graphs to further improve the segmentation performance.}
	\label{fig:graphs2}
\end{figure*}

To verify the proposed AFA-graph with other graph construction, we list the performance of different graphs in Table~\ref{different visual fea}. The setting of scale $k$ is the same as adjacency-graph~\cite{li2012segmentation}. To achieve the optimal performance of these graphs, we follow the procedure stated in~\cite{li2012segmentation} and~\cite{wang2015global} to manually select the best group in Tcut~\cite{li2012segmentation} ranking from 1 to 40. In adjacency-graph, the standard deviation of the Gaussian kernel function is defined as 20. We construct the NOLRR-graph, and the parameter $d$ is set to 50.
Our method is highly efficient and adaptive to the combination of local graph (adjacency-graph) and global graph (NOLRR-graph). It achieves the best performances in comparison with the adjacency-graph and NOLRR-graph. As shown in Fig.~\ref{fig:graphs2}, adjacency-graph only considers the local structure easily leading to wrong segmentation when the objects cover a large part of images. Our proposed NOLRR-graph often produces a dense graph, which cannot satisfy the sparsity of the desired graph. Our AFA-graph combines different affinity graphs with assimilating the advantages of two graphs to further improve the segmentation performance.

%\subsubsection{Influence of group $k_T$}
%To examine the influence of group $k_T$ on our OLRR-graph and different basic graphs, we report the indexes of PRI, VoI, GCE, and BDE for different $k_T$ ranging from 1 to 40 on BSD300 dataset. The experimental results are shown in Figs.~\ref{fig:ktinfluence}(a)-(d). From Fig.~\ref{fig:ktinfluence}, we can find that the PRI and BDE of our AFA-graph and other basic graphs become stable as the $k_T$ increases to 10. The GCE reaches its maximum when $k_T$ is 10, and then it decreases to be steady gradually. But the VoI gradually increases with increase of $k_T$. Furthermore, both adjacency-graph and LRR-graph outperform $\ell_0$-graph, $\ell_1$-graph, and $\ell_2$-graph. Our OLRR-graph achieves the best performance. To further explore the influence of group $k_T$ on our method, we show visual results of various $k_T$ ($k_T$=2, 5, 10, 20, 30, and 40) in Fig.~\ref{fig:ktinfluenceof-afagraph}. The results show that visually meaningful segmentation can be obtained by the carefully tuning of the $k_T$. When the $k_T$ is increased, our AFA-graph enforces the global structure over the superpixels and masters the meaning of regions. Moreover, our method preserves more local information in the superpixels.

\subsubsection{Different modules}
To see how AFA-graph is affected by different modules, we report the scores of combinations of different modules using in the baseline. The baseline easily combines adjacency-graph and NOLRR-graph by element-wise sum operation. The Area means the node selection method from GL-graph\cite{wang2015global}. The SPR and APC are utilized to select global nodes. Especially if the baseline does not select APC, we will use \emph{k}-means with $k=2$ for fair comparison. The results of different modules using in the baseline are shown in Table~\ref{different modules}. The results show that the segmentation performance of APC with SPR is better than area and \emph{k}-means. Therefore, the SPR can better represent superpixels compared with the area. Obviously, our method is suitable for the selection of global nodes when we know nothing about the feature distribution of the superpixels beforehand.

\begin{table}[!t]
\centering
\renewcommand{\arraystretch}{1.3}
\caption{Quantitative comparison for different methods using in our AFA-graph on BSD300 dataset. The baseline easily combines adjacency-graph and NOLRR-graph by element-wise sum operation.}
\label{different modules}
\begin{tabular}{lcccccccccc}
\toprule
Methods & PRI $\uparrow$  & VoI $\downarrow$ & GCE $\downarrow$ & BDE $\downarrow$     \\
\midrule
Baseline                         & 0.80  & 2.46  & 0.26 & 15.96    \\
Baseline+IKDE                    & 0.80  & 2.40  & 0.27 & 15.60    \\
Baseline+IKDE+Area               & 0.84  & 1.69  & 0.21 & 14.95    \\
Baseline+IKDE+\emph{k}-means     & 0.84  & 1.69  & 0.19 & 15.01    \\
Baseline+IKDE+\emph{k}-means+SPR & 0.84  & 1.68  & 0.19 & 14.94    \\
Baseline+IKDE+APC+SPR            & 0.84  & 1.67  & 0.19 & 14.72    \\
\bottomrule
\end{tabular}
\end{table}

\begin{table}[!t]
	\centering
	\renewcommand{\arraystretch}{1.3}
	\caption{Quantitative results of the proposed framework with state-of-the-art approaches on BSD300 dataset. We directly take their evaluations reported in publications for fair comparison. The 'C' in class means clustering-based methods. The 'G' in category means graph-based methods.}
	\label{tab:BSDS300}
	\begin{tabular}{lccccc}
		\toprule
		Methods  &Class & $\textrm{PRI}\uparrow$    & $\textrm{VoI}\downarrow$   & $\textrm{GCE}\downarrow$   & $\textrm{BDE}\downarrow$  \\
		\midrule
		FCM~\cite{Lei2018fuzzy}               & C & 0.74 & 2.87 & 0.41 & 13.78  \\
		FRFCM~\cite{8265186}                  & C & 0.75 & 2.62 & 0.36 & 12.87  \\
		MS~\cite{comaniciu2002mean}           & C & 0.76 & 2.48 & 0.26 & 9.70   \\
		SFFCM~\cite{Lei2018fuzzy}             & C & 0.78 & 2.02 & 0.26 & 12.90  \\
		FCR~\cite{4480125}                    & C & 0.79 & 2.30 & 0.21 & \textbf{8.99}  \\
		H\_+R\_Better~\cite{li2018iterative}  & C & 0.81 & 1.83 & 0.21 & 12.16  \\
		Corr-Cluster~\cite{Kim2013Task}       & C & 0.81 & 1.83 & --   & 11.19  \\
		HO-CC~\cite{nowozin2014image}         & C & 0.81 & 1.74 & --   & 10.38  \\     
		\midrule        
		FH~\cite{Felzenszwalb2004Efficient}   & G & 0.71 & 3.39 & 0.17 & 16.67  \\
		Ncut~\cite{shi2000normalized}         & G & 0.72 & 2.91 & 0.22 & 17.15  \\		
		MNCut~\cite{cour2005spectral}         & G & 0.76 & 2.47 & 0.19 & 15.10  \\
		LFPA~\cite{Tae2013Learning}           & G & 0.81 & 1.85 & 0.18 & 12.21 \\
		SAS~\cite{li2012segmentation}         & G & 0.83 & 1.68 & 0.18 & 11.29  \\
		$\ell_0$-Graph~\cite{wang2013graph}   & G & 0.84 & 1.99 & 0.23 & 11.19  \\
		GL-Graph~\cite{wang2015global}        & G & 0.84 & 1.80 & 0.19 & 10.66  \\
		AASP-Graph~\cite{zhang2019aaspgraph}  & G & 0.84 & 1.65 & \textbf{0.17} & 14.64 \\	
		AFA-Graph (Gaussian)                  & G & 0.84 & 1.76 & 0.20 & 15.71 \\
		AFA-Graph (Bilateral)                 & G & 0.84 & 1.68 & 0.18 & 15.26 \\	
		AFA-Graph (IKDE)                      & G & \textbf{0.84} & \textbf{1.65} & 0.18 & 15.00 \\
		\bottomrule
	\end{tabular}
\end{table}

\begin{table}[!t]
	\centering
	\renewcommand{\arraystretch}{1.3}
	\caption{Quantitative results of the proposed framework with state-of-the-art approaches on BSD500 dataset. We directly take their evaluations reported in publications for fair comparison.}
	\label{tab:BSDS500}
	\begin{tabular}{lccccc}
		\toprule
		Methods   &Class & $\textrm{PRI}\uparrow$    & $\textrm{VoI}\downarrow$   & $\textrm{GCE}\downarrow$   & $\textrm{BDE}\downarrow$  \\
		\midrule
		DSFCM~\cite{8543645}                  & C & 0.74 & 2.90 & 0.41 & --     \\
		FCM~\cite{Lei2018fuzzy}               & C & 0.74 & 2.88 & 0.40 & 13.48  \\
		MSFCM~\cite{9120181}                  & C & 0.74 & 2.85 & 0.40 & --     \\
		FRFCM~\cite{8265186}                  & C & 0.76 & 2.67 & 0.37 & 12.35  \\
		HS~\cite{Wu20}                        & C & 0.76 & 2.39 & 0.26 & 14.03  \\
		HO-CC~\cite{nowozin2014image}         & C & 0.83 & 1.79 & --   & \textbf{9.77}  \\
		AFC~\cite{8770118}                    & C & 0.76 & 2.05 & 0.22 & 12.95  \\
		RSFFC~\cite{9162644}                  & C & 0.78 & 2.12 & 0.28 & --     \\
		SFFCM~\cite{Lei2018fuzzy}             & C & 0.78 & 2.06 & 0.26 & 12.80  \\
		MS~\cite{comaniciu2002mean}           & C & 0.79 & 1.85 & 0.26 & --     \\
		\midrule
		MNCut~\cite{cour2005spectral}         & G & 0.76 & 2.33 & --   & --     \\
		FH~\cite{Felzenszwalb2004Efficient}   & G & 0.79 & 2.16 & --   & --     \\
		SAS~\cite{li2012segmentation}         & G & 0.80 & 1.92 & --   & --     \\
		RAG~\cite{7484679}                    & G & 0.81 & 1.98 & --   & --     \\
		$\ell_0$-Graph~\cite{wang2013graph}   & G & 0.84 & 2.08 & 0.23 & 11.07  \\
		AASP-Graph~\cite{zhang2019aaspgraph}  & G & 0.84 & 1.71 & \textbf{0.18} & 13.78 \\
		AFA-Graph (Gaussian)                  & G & 0.83 & 1.80 & 0.20 & 15.18 \\
		AFA-Graph (Bilateral)                 & G & 0.84 & 1.72 & 0.19 & 14.41 \\	
		AFA-Graph (IKDE)                      & G & \textbf{0.84} & \textbf{1.70} & 0.19 & 14.14 \\
		\bottomrule
	\end{tabular}
\end{table}

\subsection{Comparison with state-of-the-art methods}
We also report quantitative comparisons with the state-of-the-art methods. The comparison results on BSD300, BSD500, MSRC, SBD, and PASCAL VOC datasets are shown in Table~\ref{tab:BSDS300} to Table~\ref{tab:VOC}, respectively. We directly take their evaluations reported in publications for fair comparison.
It can be noticed that our AFA-graph has a better performance compared with the state-of-the-art methods. We attribute this to adaptive combination of different graphs, which helps to accurately separate the foreground and background. 

Our method follows a similar, but not identical strategy as the SAS, $\ell_0$-graph, and GL-graph. Instead of using only adjacent neighborhoods of superpixels in SAS and $\ell_0$ affinity graph of superpixels in $\ell_0$-graph algorithm, we build an AFA-graph by combining NOLRR-graph and adjacency-graph, which allows the constructed graph to have the characteristics of a long-range neighborhood topology, with the sparsity and high discriminating power.
The main differences between GL-graph and our method are global node selection and graph construction. In GL-graph, the superpixels are simply divided into three parts: small, medium, and large sized according to their area. For small and large sets, all superpixels connect to their adjacent neighbors, while for medium set, the superpixels are used to build a $\ell_0$-graph. However, according to the analysis of SPR and area revealed in Section~\ref{node_selection}, the area does not accurately reflect the distribution of global nodes in superpixels. In our method, the adjacency-graph of all superpixels is updated by the NOLRR-graph, which is built through NOLRR with the global nodes selected by the SPR and APC.

\begin{figure*}[!t]
	\centering
	\includegraphics[width=7in]{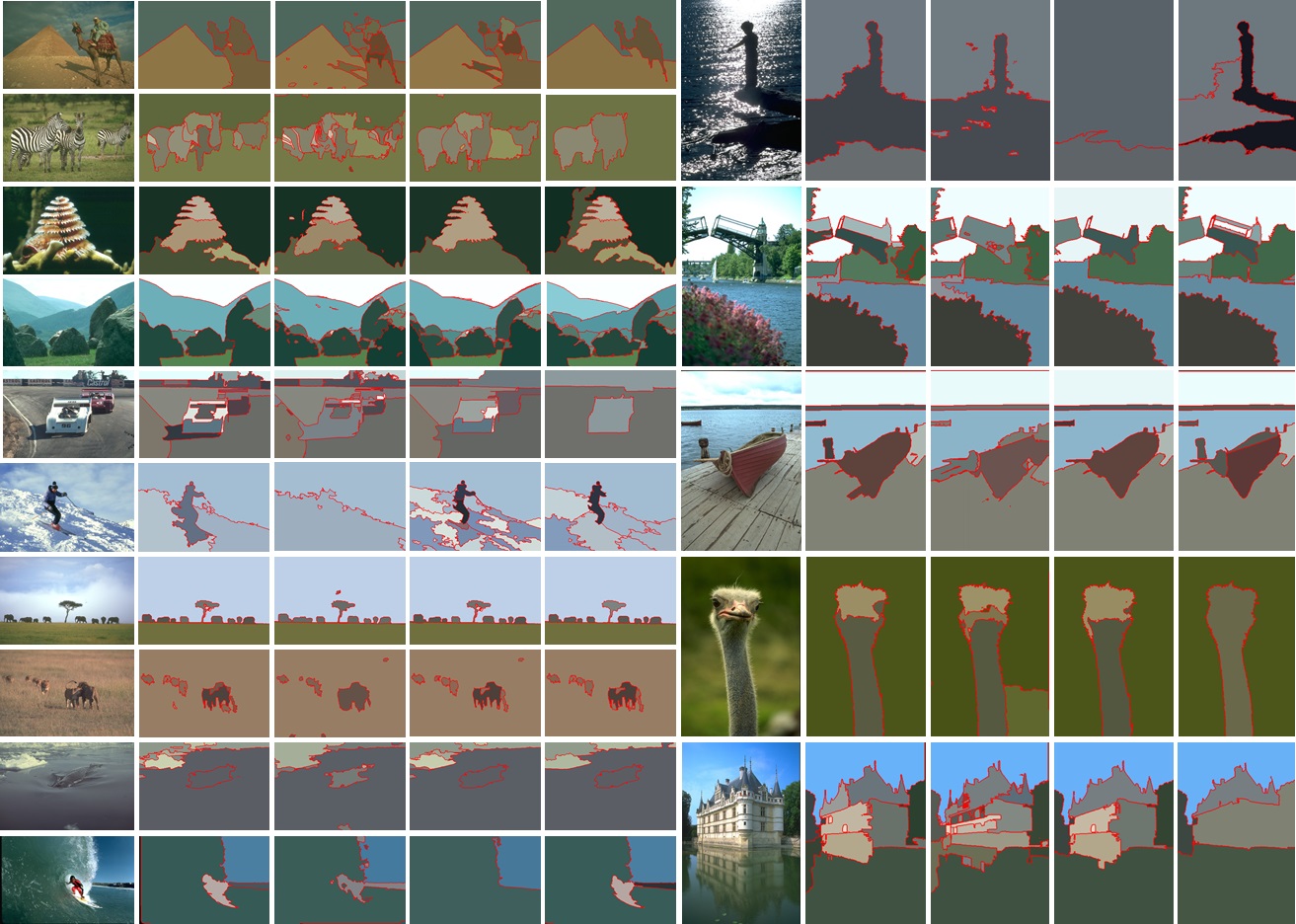}
	\caption{Visual comparison on BSD300 dataset obtained with the adjacency-graph~\cite{li2012segmentation} , $\ell_0$-graph~\cite{wang2013graph}, GL-graph~\cite{wang2013graph}, and our AFA-graph. Two columns of the comparison results are shown here. From left to right, input images, the results of the SAS, $\ell_0$-graph, GL-graph, and AFA-graph are presented. The results of AFA-graph are visually better, in particular often more accurate.}
	\label{compsas}
\end{figure*}

\begin{figure}[!t]
\centering
\includegraphics[width=3.3in]{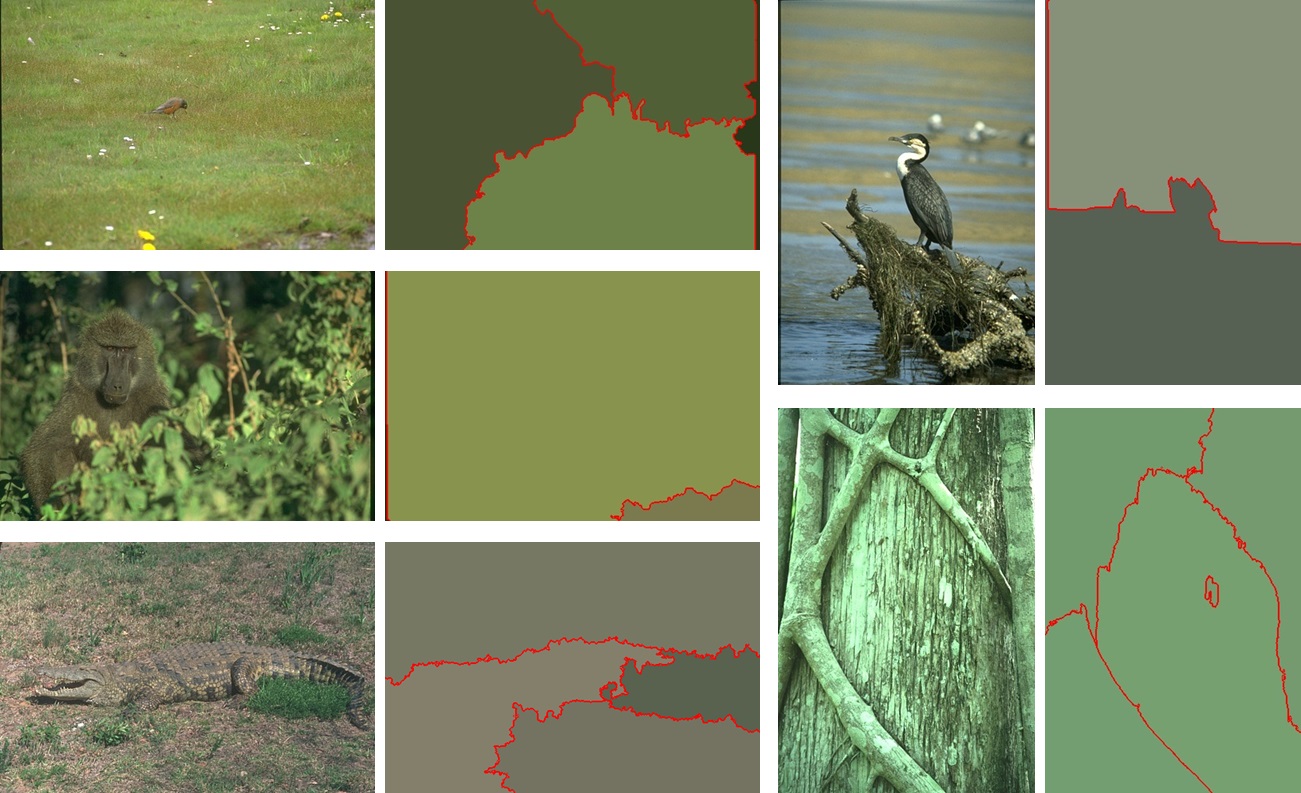}
\caption{False results by AFA-graph. Our AFA-graph only uses pixel color information, which fails to capture enough contour and texture cues of segmenting images.}
\label{fig:falseressult}
\end{figure}

Figure~\ref{compsas} shows various segmentation results obtained with SAS, $\ell_0$-graph, GL-graph, and our AFA-graph. The results of other three methods are the best results reported by the authors. It can be seen that our method can achieve a desirable result with less tuning for the group $k_{T}$ in Tcut (for ostrich, the group $k_{T}=2$), because the method takes into account the global information exactly. Especially, compared with the other three methods, our method obtains an accurate segmentation even in the difficult cases where: i) the detected object is highly textured, and the background may be highly unstructured, such as coral, surfer, and skier; ii) objects of the same type appear in a large and fractured area of the image, such as ostrich, zebras, and racecars.
However, the BDE of our method is unsatisfied. As shown in Fig.~\ref{fig:falseressult}, our method can not obtain accurate and clear boundaries, when the detected object is too tiny and its texture is easily to be confused with background. The main reason is that our AFA-graph only uses pixel color information, which fails to capture enough contour and texture cues of segmenting images.

\begin{table}[!t]
	\centering
	\renewcommand{\arraystretch}{1.3}
	\caption{Quantitative results of the proposed framework with state-of-the-art approaches on MSRC dataset. We directly take their evaluations reported in publications for fair comparison.}
	\label{tab:MSRC}
	\begin{tabular}{lccccc}
		\toprule
		Methods  &Class & $\textrm{PRI}\uparrow$    & $\textrm{VoI}\downarrow$   & $\textrm{GCE}\downarrow$   & $\textrm{BDE}\downarrow$  \\
		\midrule
		MSFCM~\cite{9120181}                  & C & 0.68  & 1.80  & 0.30   & --     \\
		DSFCM~\cite{8543645}                  & C & 0.69  & 1.91  & 0.32   & --     \\
		FCM~\cite{Lei2018fuzzy}               & C & 0.70  & 1.93  & 0.32   & 12.67  \\
		FRFCM~\cite{8265186}                  & C & 0.71  & 1.79  & 0.30   & 12.23  \\
		SFFCM~\cite{Lei2018fuzzy}             & C & 0.73  & 1.58  & 0.25   & 12.49  \\
		RSFFC~\cite{9162644}                  & C & 0.75  & 1.51  & 0.24   & --     \\
		Corr-Cluster~\cite{Kim2013Task}       & C & 0.77  & 1.65  & --     & 9.19   \\
		HO-CC~\cite{nowozin2014image}         & C & 0.78  & 1.59  & --     & \textbf{9.04}   \\	
		\midrule
		Supervised-NCut~\cite{Kim2013Task}    & G & 0.60  & 3.10  & --     & 13.50  \\
		MNCut~\cite{cour2005spectral}         & G & 0.63  & 2.77  & --     & 11.94  \\
		SAS~\cite{li2012segmentation}         & G & 0.80  & 1.39  & --     & --     \\
		%Ncut~\cite{shi2000normalized}         & G & 0.81  & \textbf{1.25}  & --     & --     \\
		$\ell_0$-Graph~\cite{wang2013graph}   & G & 0.82  & 1.29  & 0.15   & 9.36   \\
		AASP-Graph~\cite{zhang2019aaspgraph}  & G & 0.82  & 1.32  & 0.14   & 13.38  \\
		AFA-Graph (Gaussian)                  & G & 0.80  & 1.38  & 0.15   & 15.85 \\
		AFA-Graph (Bilateral)                 & G & 0.81  & 1.32  & 0.14   & 13.92 \\	
		AFA-Graph (IKDE)                      & G & \textbf{0.82} & \textbf{1.29} & \textbf{0.14} & 13.37 \\
		\bottomrule
	\end{tabular}
\end{table}

\subsection{Time complexity analysis}
Our method includes steps of over-segmentation, feature extraction, global node selection, NOLRR-graph construction, and Tcut. The time complexity of OMP~\cite{you2016scalable}  global node selection and NOLRR-graph construction are analyzed in Section~\ref{node_selection} and Section~\ref{graph_construction}, respectively. The time complexity of Tcut is $O(k_T|N|^{3/2})$ with a small constant. Using an aforementioned computer, our method takes totally 7.61 seconds to segment an image with size 481$\times$321 from BSD300 on average, where 5.11 seconds for generating superpixels, 0.56 seconds for global node selection, 1.12 seconds for building adaptive affinity graph, as well as only 0.82 seconds for the bipartite graph construction and partitioning with Tcut. The average running times (ARTs) of methods are obtained on aforementioned computer. The ART of the proposed AFA-graph is close to SAS and slight faster than GL-graph and our previous AASP-graph. Note that the AFA-graph is more efficient than GL-graph and AASP-graph in regard to adaptive node selection and their graph construction. In contrast, $\ell_0$-graph, MNcut, GL-graph, and Ncut usually take more than 20, 30, 100, and 150 seconds, respectively. The main reason is that extracting different features cost too much time.

In summary, there are four main reasons why our AFA-graph has high performance: i) we propose IKDE to estimate the mLab of natural images to reduce the noise; ii) the combination of SPR and APC can select the global nodes, accurately mining the feature distribution of superpixels; iii) the NOLRR-graph is proposed to reduce time complexity while improving segmentation accuracy; iv) our AFA-graph combines adjacency-graph and NOLRR-graph with assimilating the advantages of two graphs.

%\begin{figure*}[!t]
%\centering
%\includegraphics[width=6.6in]{figures/8.jpg}
%\caption{Visual segmentation examples by AFA-graph on BSD dataset.}
%\label{fig:visualressultbsd}
%\end{figure*}

%\begin{figure*}[!t]
%\centering
%\includegraphics[width=6.6in]{figures/9.jpg}
%\caption{Visual segmentation examples by AFA-graph on MSRC dataset.}
%\label{fig:visualressultmsrc}
%\end{figure*}

\begin{table}[!t]
	\centering
	\renewcommand{\arraystretch}{1.3}
	\caption{Quantitative results of the proposed framework with state-of-the-art approaches on SBD dataset.}
	\label{tab:SBD}
	\begin{tabular}{lccccc}
		\toprule
		Methods  &Class & $\textrm{PRI}\uparrow$    & $\textrm{VoI}\downarrow$   & $\textrm{GCE}\downarrow$   & $\textrm{BDE}\downarrow$  \\
		\midrule
		SFFCM~\cite{Lei2018fuzzy}            & C & 0.66  & 1.89  & 0.22   & 16.44  \\
		\midrule
		$\ell_0$-Graph~\cite{wang2013graph}  & G & 0.80  & 1.97  & 0.20   & \textbf{9.96}   \\
		AASP-Graph~\cite{zhang2019aaspgraph} & G & 0.81  & 1.78  & \textbf{0.17}   & 10.61  \\
		SAS~\cite{li2012segmentation}        & G & 0.81  & 1.78  & 0.17   & 10.52  \\
		AFA-Graph (Gaussian)                 & G & 0.80  & 1.78  & 0.20   & 11.11 \\
		AFA-Graph (Bilateral)                & G & 0.81  & 1.75  & 0.18   & 10.67 \\	
		AFA-Graph (IKDE)                     & G & \textbf{0.81} & \textbf{1.75} & 0.18 & 10.33 \\
		\bottomrule
	\end{tabular}
\end{table}

\begin{table}[!t]
	\centering
	\renewcommand{\arraystretch}{1.3}
	\caption{Quantitative results of the proposed framework with state-of-the-art approaches on PASCAL VOC dataset.}
	\label{tab:VOC}
	\begin{tabular}{lccccc}
		\toprule
		Methods  &Class & $\textrm{PRI}\uparrow$    & $\textrm{VoI}\downarrow$   & $\textrm{GCE}\downarrow$   & $\textrm{BDE}\downarrow$  \\
		\midrule
		SFFCM~\cite{Lei2018fuzzy}            & C & 0.61  & \textbf{1.38}  & 0.21   & 37.69  \\
		\midrule
		AASP-Graph~\cite{zhang2019aaspgraph} & G & 0.58  & 1.83  & 0.18   & 41.94  \\
		SAS~\cite{li2012segmentation}        & G & 0.61  & 1.65  & 0.17   & 39.13  \\
		$\ell_0$-Graph~\cite{wang2013graph}  & G & 0.61  & 1.59  & 0.17   & 38.73  \\
		AFA-Graph (Gaussian)                 & G & 0.61  & 1.61  & 0.17   & 35.42 \\
		AFA-Graph (Bilateral)                & G & 0.62  & 1.56  & 0.16   & 34.14 \\	
		AFA-Graph (IKDE)                     & G & \textbf{0.62} & 1.55 & \textbf{0.16} & \textbf{34.47} \\
		\bottomrule
	\end{tabular}
\end{table}

%% file: files/CONCLUSION.tex
\section{CONCLUSION}
\label{sec:con}
In this paper, an AFA-graph is proposed to obtain good results for natural image segmentation. The method uses superpixels of different scales as segmentation primitive. To reduce the noise, we then introduce an IKDE method to estimate the mLab features of superpixels. Besides, the SPR and APC are applied to select global sets, which are used to build a NOLRR-graph to update an adjacency-graph of superpixels at each scale. Experimental results show the good performance and high efficiency of the proposed AFA-graph. We also compare our AFA-graph with the state-of-the-art methods, and our AFA-graph achieves competitive results on five benchmark datasets namely BSD300, BSD500, MSRC, SBD, and PASCAL VOC.

In the future, we plan to extend the proposed framework to include more discriminating feature such as texture, shape, and priors to obtain better results. And we
will explore the combination of graph and deep unsupervised learning to improve image segmentation performance. 